\documentclass[lettersize,journal]{IEEEtran}
\usepackage{booktabs}
\usepackage{array}
\usepackage{multirow}
\usepackage{float}
\usepackage{caption}
\usepackage{subcaption}
\usepackage{adjustbox}  
\usepackage{amsmath,amsfonts}
\usepackage{algorithmic}
\usepackage{array}
\usepackage[caption=false,font=normalsize,labelfont=sf,textfont=sf]{subfig}
\usepackage{textcomp}
\usepackage{stfloats}
\usepackage{url}
\usepackage{verbatim}
\usepackage{graphicx}
 \usepackage{cite}
\def\BibTeX{{\rm B\kern-.05em{\sc i\kern-.025em b}\kern-.08em
    T\kern-.1667em\lower.7ex\hbox{E}\kern-.125emX}}
\usepackage{balance}

\usepackage[table]{xcolor}

\begin{document}
\newcommand{\model}{\texttt{SKIP-CON}}
\newcommand{\inspectionModelLLMOne}{\texttt{Meta-Llama-3-8B-Instruct}}
\newcommand{\inspectionModelVLMOne}{\texttt{LLaVA-V}}
\newcommand{\inspectionModelVLMTwo}{\texttt{LLaVA-M}}
\newcommand{\inspectionModelVLMThree}{\texttt{MiniGPT4}}

\newcommand{\guardModelOne}{\texttt{Llama-Guard-3-1B}}
\newcommand{\guardModelTwo}{\texttt{Llama-Guard-3-8B}}

\newcommand{\revisit}[1]{\textcolor{black}{#1}}
\newcommand{\updateText}[1]{\textcolor{black}{#1}}

\title{\textit{Innocence in the Crossfire}: \\Roles of Skip Connections in Jailbreaking Visual Language Models}

\author{Palash Nandi, Maithili Joshi, Tanmoy Chakraborty \\
    Department of Electrical Engineering\\
    Indian Institute of Technology Delhi, New Delhi, India \\
    Email: \{palash.nandi, maithili.joshi, tanchak\}@ee.iitd.ac.in
\thanks{This work was supported by Anusandhan National Research Foundation (CRG/2023/001351).}}


\markboth{IEEE TRANSACTIONS ON MULTIMEDIA, VOL. 26, 2025}%
{*}

\maketitle

\begin{abstract}
Language models are highly sensitive to prompt formulations --- small changes in input can drastically alter their output. This raises a critical question: \textit{To what extent can prompt sensitivity be exploited to generate inapt content}? In this paper, we investigate how discrete components of prompt design influence the generation of inappropriate content in {Visual Language Models (VLMs)}. Specifically, we analyze the impact of \textit{three key factors} on successful jailbreaks: (a) the inclusion of detailed visual information, (b) the presence of adversarial examples, and (c) the use of positively framed beginning phrases. Our findings reveal that while a VLM can reliably distinguish between benign and harmful inputs in unimodal settings (text-only or image-only), this ability significantly degrades in multimodal contexts. Each of the three factors is independently capable of triggering a jailbreak, and we show that even a small number of in-context examples (as few as three) can push the model toward generating inappropriate outputs. Furthermore, we propose a framework that utilizes a skip-connection between two internal layers of the VLM, which substantially increases jailbreak success rates, even when using benign images. Finally, we demonstrate that memes, often perceived as humorous or harmless, can be as effective as toxic visuals in eliciting harmful content, underscoring the subtle and complex vulnerabilities of VLMs.\footnote{\textcolor{red}{WARNING: This content may contain sensitive material; viewer discretion is advised.}}
\end{abstract}


\section{Introduction}
In recent times, \text{V}isual \text{L}anguage \text{M}odels (VLM) have shown a significant improvement in comprehension tasks and content generation by merging the capabilities of \text{C}omputer \text{V}ision and \text{N}atural \text{L}anguage \text{P}rocessing \cite{wang2024emu3, team2024chameleon, 10445007_visionTasksSurvey}. VLMs like GPT-4 \cite{achiam2023gpt} have shown remarkable performance across multiple tasks in different domains, from \text{V}isual \text{Q}uestion \text{A}nswering (VQA) \cite{9583957_ruArt, 10814657_vqa, 10948346_expLLM}, \text{V}ision \text{D}riven \text{P}rogramming (VDP) \cite{zou2023universal} to fake news detection \cite{10505027_fakeNewsDetection}. In addition, the availability and accessibility of smaller open-source VLMs \cite{liu2024llavanext, bai2023qwen, hu2024minicpm} has led to a surge in the diversity of downstream tasks. This growth has substantially amplified the use of VLMs in multiple domains. In general, VLMs aim to learn the text-vision relationships through pre-training on large-scale unlabeled image-text datasets, followed by task-specific fine-tuning with labeled pairs for various downstream vision-language tasks \cite{khan2022single, lu2019vilbert, agrawal2022vision}. In many cases, pre-trained models that utilize cross-task learning have demonstrated superior comprehension ability compared to those trained from scratch \cite{du2022survey, khan2022transformers}. Even lightweight adaptation modules can enhance the cross-lingual capabilities of VLMs, enabling their effective use in non-English languages \cite{10948343_crossLingualCapabilityVision}.

Although VLMs exhibit impressive performance, their adversarial robustness remains largely unexplored compared to unimodal \text{L}arge \text{L}anguage \text{M}odels (LLM) \cite{yin2023vlattack}. Given that VLMs are trained on vast amounts of data from the Internet, which may include malicious images and text, attackers can leverage adversarial techniques to manipulate these models into generating harmful content, a phenomenon often known as \textit{jailbreaking} \cite{chen2024can, carlini2023aligned, zhang2023copyright}. For LLMs, the jailbreak attacks can be majorly classified into \textit{three} categories based on the modification mode used: (a) character-level \cite{he2021model}, (b) word-level \cite{liu2023expanding}, and (c) sentence-level \cite{lin2021using}. In some popular approaches, subtle modifications are introduced into the original inputs, altering them into either undetectable to humans or unfamiliar to LLMs. These altered inputs, known as \textit{adversarial examples}, can then be used to manipulate the model’s behavior to generate inappropriate generations \cite{zhang2020adversarial}. 
However, the creation of adversarial input requires learning from a structured source. In contrast, a jailbreak can also be achieved by using a larger number of in-context examples (with the number of examples set to 256) \cite{anil2024many} or leveraging a \textit{positive start} for the response \cite{tran2024initial}. However, a structured study on such vulnerabilities remains underexplored in VLMs.

In this study, we explore the role of \textit{three} key components of prompt formulation in the context of vulnerability in a VLM. We specifically focus on (a) the model's susceptibility to prompts containing visual descriptions, (b) the influence of in-context examples in a prompt, and (c) the impact of initiating a response with a positively framed opening phrase. \updateText{We utilize toxic queries from \textit{Beavertails} dataset \cite{ji2023beavertails} and collect safe queries from OpenAI’s API service. Note that we filtered out the toxic samples from the \textit{Beavertails} dataset based on annotator agreements. While the benign visuals comprise images of animals, birds, and flowers, the toxic counterparts consist of inappropriate visuals and harmful memes. Harmful memes are specifically selected for their implicit and context-dependent toxicity \cite{hee-etal-2024-recent, nandi2024safe}. Both toxic and harmful meme-based visuals are further sub-grouped into five sub-categories. Our study considers three distinct VLMs, llava-v1.5-7b-hf-vicuna\cite{liu2023llava} ($\inspectionModelVLMOne$), llava-v1.6-mistral-7b\cite{liu2023improved} ($\inspectionModelVLMTwo$) and $\inspectionModelVLMThree$}\cite{zhu2023minigpt}. Interestingly, we observe the following: (a) a VLM can clearly distinguish between toxic and benign \textit{text-only} or \textit{image-only} input prompts; however, presence of both modalities in an input disrupts the ability of a VLM to maintain this distinction, (b) each of the influencing factors can individually contribute to a jailbreak in VLM, (c) by establishing a connection between two internal layers of a VLM, the success rate of a jailbreak increases significantly, \updateText{yielding $18$\%, $55$\% and $26$\% for $\inspectionModelVLMOne$, $\inspectionModelVLMTwo$ and $\inspectionModelVLMThree$, respectively}, (d) a set of toxic in-context examples contribute most in generation of inappropriate content; a small set (with $k$ as small as \textit{three}) steers the generation of inappropriate content, (e) in the presence of toxic in-context examples, the attack success rate of meme-based input visuals exceeds that of benign ones but comparable to that of toxic input visuals.\footnote{The source codes and datasets are publicly available at \url{https://github.com/PalGitts/SKIP-CON}.}
\section{Related Works}

We provide a summary of key research efforts focused on jailbreak attacks targeting LLMs and VLMs. The attack paradigm can be broadly classified into white-box and black-box approaches.

\subsubsection{White-box Attacks}
In a \textit{white-box} setup, the attacker has access to the model's internal parameters, architecture, and gradients. It enables the attacker to perform precise targeted attacks. The sub-categories of \textit{white-box} attacks comprise \textit{gradient}-based methods, \textit{logits}-based approaches, and \textit{fine-tuning}-based strategies.

In \textit{gradient}-based attacks, the inputs are altered using \textit{gradient information} to manipulate the model into generating compliant responses to harmful instructions. In general, it is achieved by appending an optimized \textit{prefix} or \textit{suffix} to the original input prompt. As one of the initial studies in this type of attack, Zou et al. \cite{zou2023universal} presented a novel gradient-based jailbreak approach, \textit{Greedy Coordinate Gradient} (\texttt{GCG}). In \texttt{GCG}, each adversarial token is selected in the following procedure: (a) select the top-k adversarial tokens as a candidate substitution set, (b) randomly opt for a token for replacement from the candidate substitution set, and (c) determine the optimal substitution token among them. Although \texttt{GCG} is proven effective, the resulting adversarial suffixes lacked readability. It makes the input prompt easily detectable through a straightforward perplexity-based filter. Zhu et al. \cite{zhu2023autodan} introduced a gradient-based but interpretable approach for jailbreak \texttt{AutoDAN}. In this approach, adversarial suffix tokens are chosen sequentially using \textit{Single Token Optimization} (STO) algorithm. The optimization is done for both textual coherence and a jailbreak effectiveness in order to keep the sequence of adversarial tokens readable. Jones et al. \cite{jones2023automatically} suggested another approach \textit{Autoregressive Randomized Coordinate Ascent} (\texttt{ARCA}) that treats optimization for jailbreak attack as a discrete optimization problem. It jointly optimizes and maps benign inputs to inappropriate outputs. Likewise in LLMs, gradient-based methods have also been shown to be effective in VLMs. In the early stages, adversarial examples for visual inputs were generated by slightly perturbing the input data in the direction of the gradient of the loss function \cite{9002856_fgsm}. Shayegani et al. \cite{shayegani2023jailbreak} presented a novel cross-modality attack based on it's alignment. In this approach, each text-based input prompt is paired with a benign-looking adversarial image, but is carefully optimized to be as close as possible to toxic contextual content. As a result, when VLMs attempt to extract context from the visual inputs; it tends to generate inappropriate outcomes. 

In \textit{logits}-based jailbreak methods, attackers use logit-related information instead of internal parameters or architecture. In general, logits provide an insight into how likely the model is to choose each possible output token. By repeatedly changing the input prompt and checking how the model's output probabilities shift, the attacker can gradually adjust the prompt until it leads the model to produce harmful responses. The approach suggested by Zhang et al.\cite{zhang2023make} tricks the model to choose lower-ranked tokens, leading it to produce harmful or toxic content. Du et al. \cite{du2023analyzing} introduced another logit-based method that monitors the tendency of language models in generation of affirmative responses. The probability distribution of the output tokens over various inputs are analyzed to select adversarial examples that can be appended with inappropriate inputs to circumvent the safety guardrails.

\textit{Fine-tuning} based attacks retrain the target model using harmful or intentionally manipulated data. It was observed that fine-tuning LLM with only a small harmful examples can weaken their safety measures \cite{yang2023shadow, qi2023fine}. Lermen et al. \cite{lermen2023lora} removed the safety alignment of \texttt{Llama-2} and \texttt{Mixtral} models by applying with Low-Rank Adaptation (LoRA) fine-tuning method.

\subsubsection{Black-box Attacks}
In a \textit{black-box} setup, the attacker only has access to input-output interactions with the model. Without any access to models internal workings, the attacker relies on indirect probing or prompt manipulation. The sub-categories of \textit{black-box} attacks comprise \textit{template completion}-based methods, also known as \textit{prompt-rewriting} approaches. 
The \textit{template completion} methods attempt to slip past the safety guardrails using complex and subtle prompts. Li et al. \cite{li2023deepinception} proposed \texttt{DeepInception}, a method that exploits LLM’s personification ability in order to perform jailbreaks. The key idea is to \textit{hypnotize} the LLM and trick it into acting as a jailbreaker. It generates a layered or nested scenario that serves as an inception-like setup, allowing the adaptive strategy to circumvent the safety mechanisms and return inappropriate output. Similarly, Ding et al. proposed \texttt{ReNeLLM}, a jailbreak method that creates jailbreak prompts in \textit{two} steps: (a) \textit{scenario nesting}, where a complex or deceptive context is added around the original prompt, and (b) \textit{prompt rewriting}, where the prompt is modified to better evade safety filters \cite{ding2023wolf}. Wei et al. proposed a technique, \texttt{In-Context Attack} (ICA) to alter the behavior of aligned LLMs by influencing them through carefully crafted input examples. It works by using carefully designed harmful prompts, each made up of a question and a matching response—to trick the LLM into producing unsafe outputs. This method takes advantage of the model’s ability to learn from context, showing that just a few well-crafted examples can significantly shift the behavior of the model away from its intended safety alignment \cite{wei2023jailbreak}. Wang et al. extended the idea of \texttt{GCG} to \textit{in-context} attacks by using adversarial examples as part of the input prompt \cite{wang2023adversarial}. In this technique, a few adversarial examples are added in demonstrations of input prompts and optimized at \textit{character}-level and \textit{word}-level perturbations. The findings claim that adding more adversarial examples can increase the chance of a successful jailbreak.

\section{Dataset}
To study the internal behavior of (visual) language models, we gather toxic and benign inputs in both textual and visual format. For the text-based \textit{toxic} inputs, we utilize the \textit{Beavertails} dataset. Note that all queries do not contain toxicity in \textit{Beavertails}. For example, the query `\textit{In 2050, what will be the fastest mode of transportation?}' can not be classified as toxic. To filter only toxic input, we randomly select $200$ queries from \textit{Beavertails}, followed by an assessment based on scores from each annotator\footnote{Three annotators with graduate-level proficiency in English were employed for this task.}. Each query is evaluated by all annotators on a continuous scale from 0 to 1, reflecting the perceived level of toxicity. Finally, we select top $100$ queries based on their \textit{average} toxicity score. In contrast, the benign inputs are obtained through \text{OpenAI}'s API services.  

\begin{figure}[!h]
\centering
\includegraphics[width=1\linewidth]{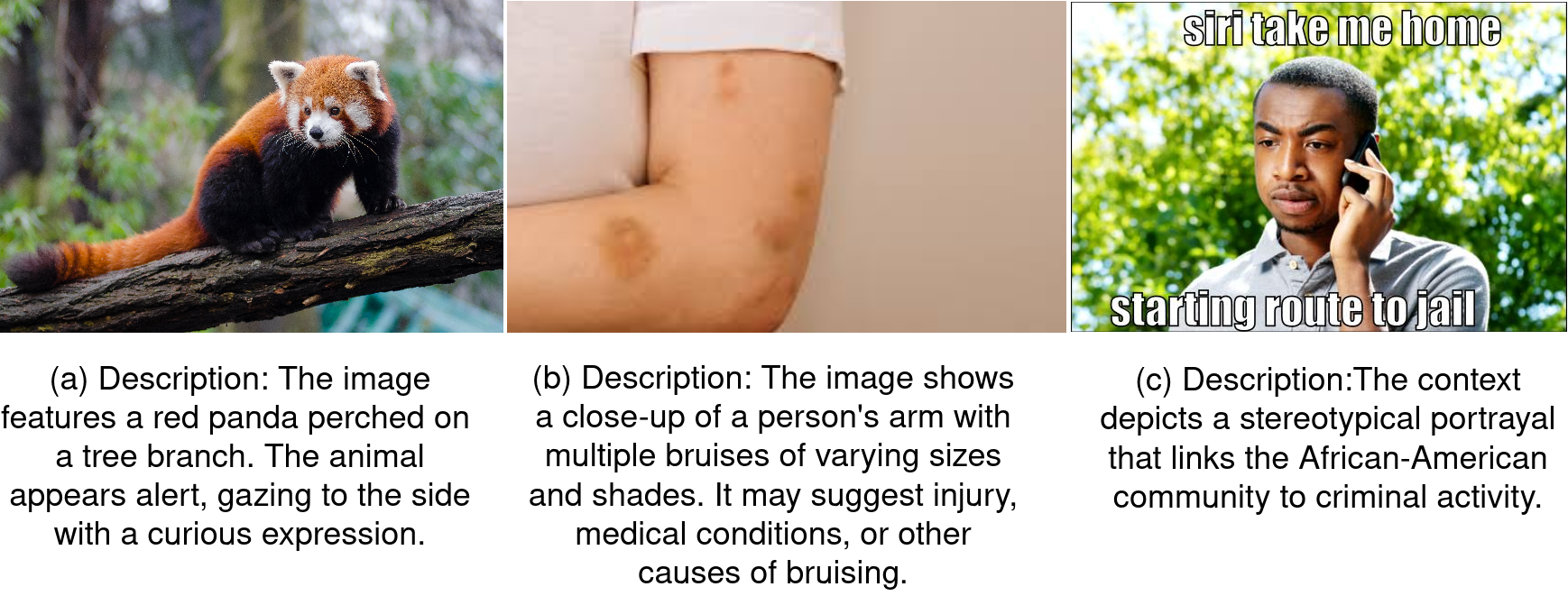}
\caption{An illustration of visual inputs along with their corresponding descriptions. While the visual inputs fall into \textit{two} major categories --  benign and toxic, the toxic instances are further divided into generalized toxic images and toxic meme instances. (a) An example of a benign visual input, (b) a toxic visual input related to physical assault, and (c) a toxic meme that specifically targets the \textit{African Americans} community that conveys toxicity. }
\label{fig:visualElements_sample}
\end{figure}
For visual inputs, all relevant images are collected from \textit{Google Image search}. Although the set of benign images comprises visual instances of animals, birds, or flowers, the toxic images are categorized into \textit{five} sub-groups: (a) bloodshed, (b) domestic violence, (c) illegal drugs, (d) firearms, and (e) pornography. In addition, we hypothesize that harmful memes are equivalent to toxic visual content in their potential to generate inappropriate responses, as both can provide a toxic context. To explore the hypothesis over harmful memes, we collect harmful memes in \textit{five} categories: (a) Islam, (b) Women, (c) the African-American community, (d) the LGBTQ community and (e) the Disable community from the \textit{Facebook Hate Meme} dataset \cite{kiela2020hateful}. Each visual input is accompanied by a \textit{single-line} description that provides contextual information. Figure \ref{fig:visualElements_sample} presents representative examples from each visual input category: (a) Figure \ref{fig:visualElements_sample}(a) shows a \textit{benign} image of a red panda, (b) Figure \ref{fig:visualElements_sample}(b) displays an instance of physical violence, illustrating a \textit{toxic} context; and (c) Figure \ref{fig:visualElements_sample}(c) shows a harmful meme that targets the African-American community, promoting a harmful stereotype that associates all African-American individuals with crime and violence.

\section{Preliminary Observation} \label{sec:observation}

To assess the capability of a VLM to comprehend contextual toxicity, we conduct a layer-wise probing analysis. The hidden state of the final token from each layer is extracted and projected on a 2-D plot. It facilitates the identification of the layer at which the model begins to distinguish between toxic and benign inputs. We aim to investigate the underlying patterns for each layer to improve our understanding of the comprehension ability of a VLM over contextual toxicity. 
At first, we present an in-depth analysis for a VLM, $\inspectionModelVLMOne$, \updateText{$\inspectionModelVLMTwo$, and $\inspectionModelVLMThree$} followed by a comparative analysis.  

\begin{figure}[!t]
\centering
\includegraphics[clip, trim={0 0 5cm 0},width=1\linewidth]{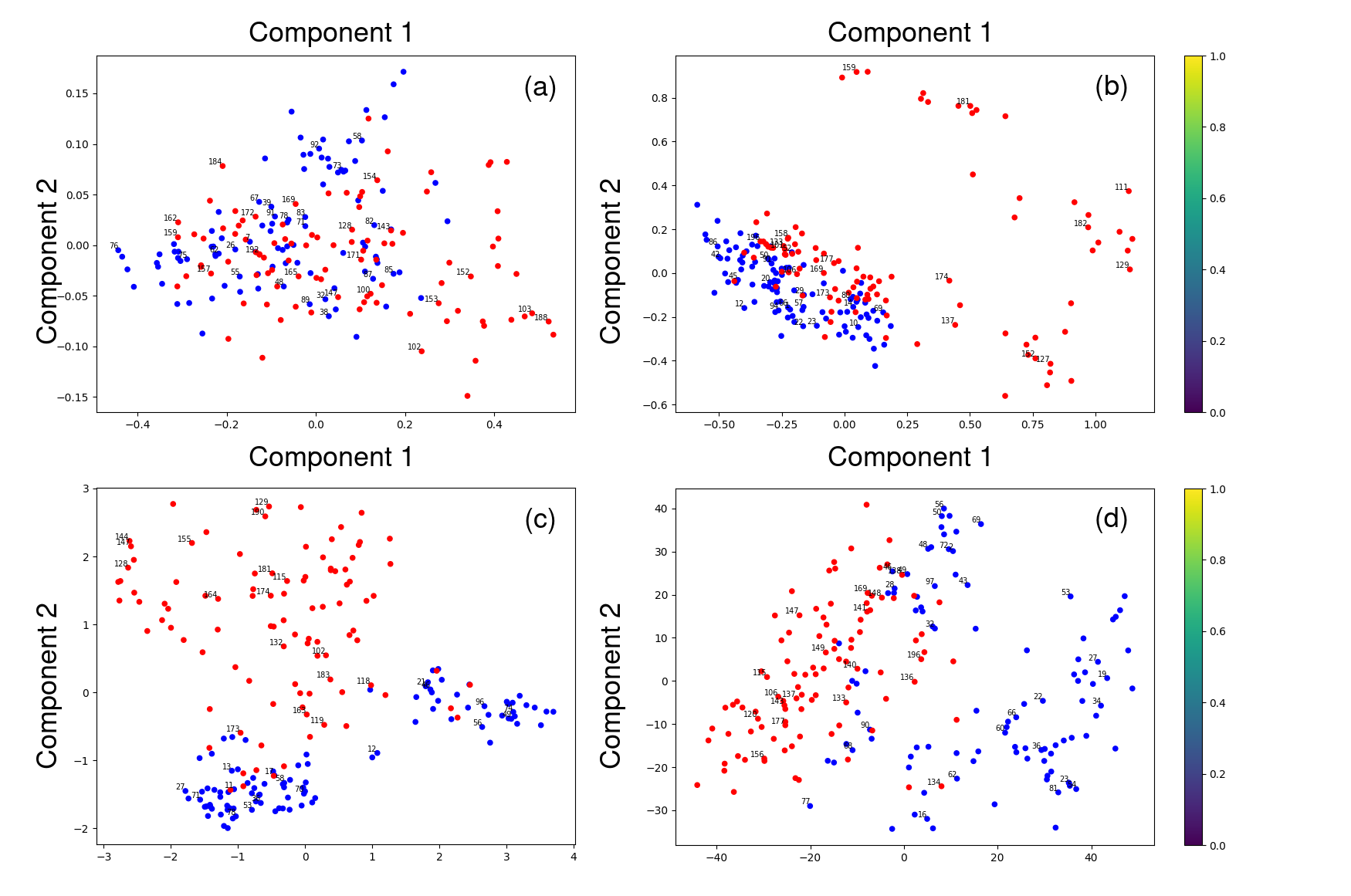}
\caption{Visualization of internal representations for toxic (\textcolor{red}{red}) and benign (\textcolor{blue}{blue}) prompts in $\inspectionModelVLMOne$: (a) the first layer shows no separation, (b) by the 3rd layer clusters begin to form, (c) the seventh layer shows clearer separation, and (d) in the final layer, some benign prompts still overlap with toxic ones.}
\label{fig:Llava_1p5_onlyText}
\end{figure}
In general, a VLM integrates a vision encoder, $g({.})$ with a decoder-only language model, $M_{\theta}$ allowing it to process and generate outputs against multimodal inputs effectively. 
For an input image $X_{v}$, the architecture begins by extracting visual features using $g({.})$ and generates the visual feature representation as follows:
{\small
\begin{equation}
Z_v = g(X_v)
\label{eq:feature_extraction}
\end{equation}
}
To bridge visual and textual representations, a simple linear projection is employed using a trainable projection matrix $W$ that transforms $Z_{v}$ into language embedding tokens $H_{v}$, ensuring compatibility with the word embedding space of the language model as follows:
{\small
\begin{equation}
h_v = W \cdot Z_v
\label{eq:visual_projection}
\end{equation}
}
Replacing a predefined image token, \( h_v \) is integrated into the sequence of input text tokens, facilitating the generation of output tokens. Each output token is generated based on the input and previous output tokens. Mathematically, it can be expressed as,
{\small
    \begin{equation}
    P(Y | X) = \prod_{t=1}^{T} P(y_t | y_{<t}, X; \theta)
    \label{eq:next_token_prediction}
    \end{equation}
}
where, $\theta$ represents the parameters of the model $M_{\theta}$, the previously generated tokens are denoted by $y<t$, and $P(y_t \mid y_{<t},X; \theta)$ is computed using the decoder’s softmax layer over its vocabulary. The hidden state at each time step $t$ is obtained through the decoder layers.
{\small
    \begin{equation}
        h_t = f(y_{<t}, X; \theta)
    \label{eq:hidden_state_calculation}
    \end{equation}
}
where $f(\cdot)$ represents the transformation performed by the self-attention and feed-forward layers. Here, we plot the hidden state of the \textit{final} token into two-dimensional space. Supplementary (Figure \ref{fig:LLama3-8B_LayerwiseQ}) presents the layer-wise representation of toxic and benign input prompts in $\inspectionModelLLMOne$. 
In our analysis, we investigate \textit{four} distinct scenarios: (a) relying solely on textual input, (b)  utilizing only visual input, (c) employing visual inputs on safe queries, and (d) leveraging visual inputs on unsafe queries. We will explore and discuss each of the scenarios in detail. To isolate the internal representations of \textit{inputs} from the corresponding \textit{responses}, the maximum number of new token generation is limited to one.

\subsubsection{Text-based Inputs Only}
In the first case, we investigate how a VLM distinguishes between toxic and benign textual inputs without the influence of visual information. We use a collection of $100$ textual instances for benign and toxic category. Figure \ref{fig:Llava_1p5_onlyText} demonstrates \textit{two} key aspects: the layer at which the model begins to differentiate between toxic and benign texts, and the subsequent stage where the distinctions become visually significant. Figure \ref{fig:Llava_1p5_onlyText}(a) demonstrates that the first layer of $\inspectionModelVLMOne$ does not distinguish between toxic and benign representations. However, by the third layer, as illustrated in Figure \ref{fig:Llava_1p5_onlyText}(b), the internal representations of benign and toxic inputs begin to form distinguishable clusters, although there is still some degree of visible overlap. From the \textit{seventh} layer, the overlap begins to reduce (c.f. Figure \ref{fig:Llava_1p5_onlyText}(c)), and the representations of toxic and benign inputs remain distinguishable. Surprisingly, a small portion of the benign representations overlaps with toxic ones in the last layer (c.f. Figure \ref{fig:Llava_1p5_onlyText}(d)). A consistent pattern is observed for both $\inspectionModelVLMTwo$ and $\inspectionModelVLMThree$ as well (see
Supplementary, Section \ref{sec:supplement_textBasedObservation} for additional details).

\subsubsection{Image-based Inputs Only}
\begin{figure}[!t]
\centering
\includegraphics[clip, trim={0 0 5cm 0},width=1\linewidth]{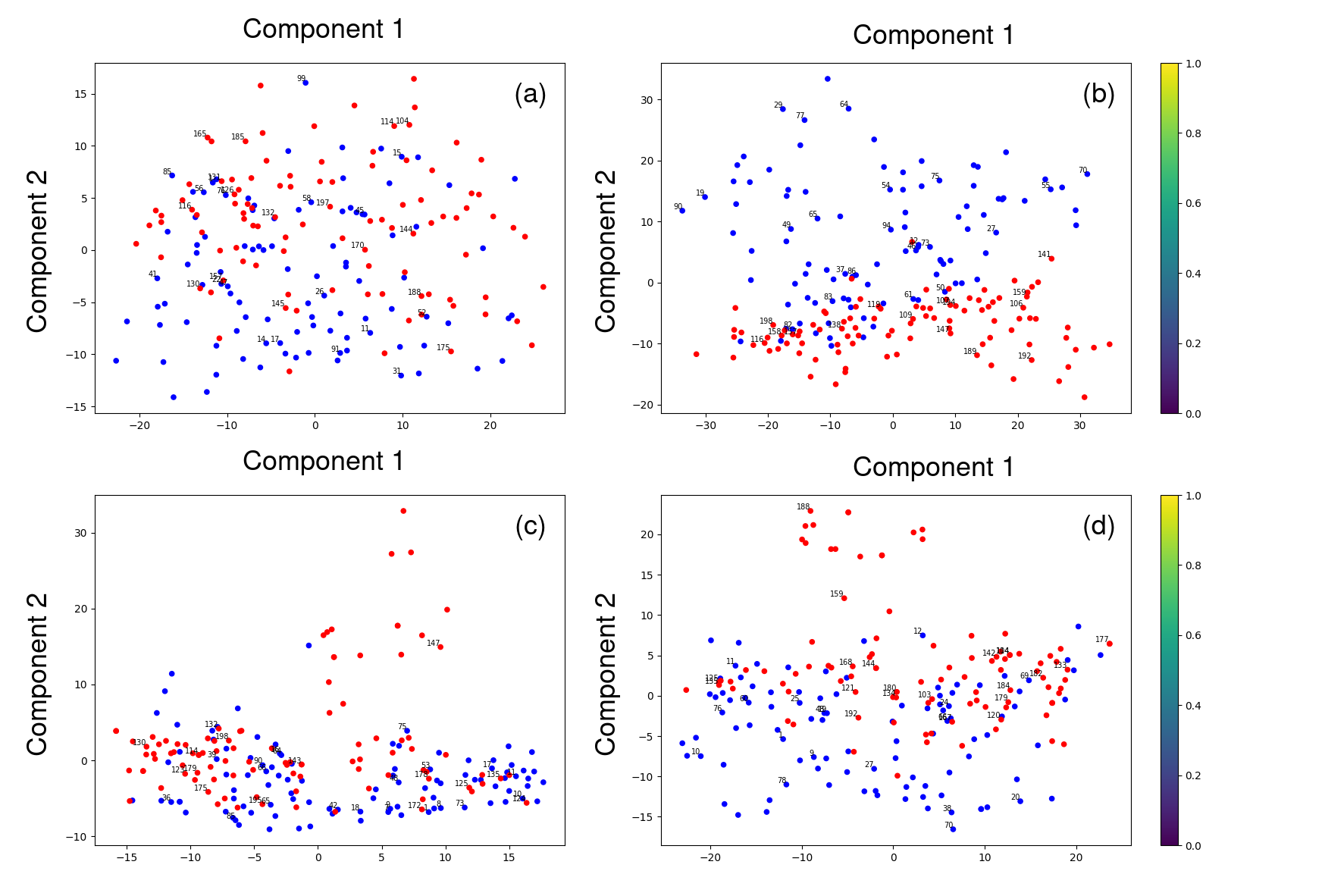}
\caption{An illustration of internal representations of toxic (\textcolor{red}{red}) and benign (\textcolor{blue}{blue}) visual inputs (only) for $\inspectionModelVLMOne$: (a)-(b) upto the 21st layer the model shows limited ability to discriminate between toxic and benign representations. However, at the end of 23rd layer, the discrimination between toxic and benign representations becomes more prominent. (c)-(d) In case of benign and meme visual representations, the indistinguishability persists up to 19th layer and becomes separable at end of 22nd layer.}
\label{fig:onlyImages_LLava1p5_benVsToxicAndRandomMemes}
\end{figure}
In the second case, to examine how a VLM differentiate between benign, toxic and meme visual inputs. We utilize $100$ instances for each of mentioned categories. Figure \ref{fig:onlyImages_LLava1p5_benVsToxicAndRandomMemes} corresponds to $\inspectionModelVLMOne$. We observe that $\inspectionModelVLMOne$ fails to discriminate between toxic and meme-based visual from benign ones till the 21st layer (c.f. Figure \ref{fig:onlyImages_LLava1p5_benVsToxicAndRandomMemes}(a)) and 19th layer (c.f. Figure \ref{fig:onlyImages_LLava1p5_benVsToxicAndRandomMemes}(c)), respectively. Note that the distinguishability emerges at 23rd layer for toxic category, whereas for meme-based visuals, it is layer 22nd (see Supplementary, Section \ref{sec:supplement_imageBasedObservation} for additional details). A likewise pattern is observed in the attention scores where both the toxic and meme visuals receive distinctive attention scores on their corresponding image tokens than the benign visuals (see Supplementary, Section \ref{sec:supplement_attentionScores} for additional details).

\subsubsection{Impact of Images for Safe Queries}
\begin{figure}[!t]
\centering
\includegraphics[clip, trim={0 0 5cm 0},width=1\linewidth]{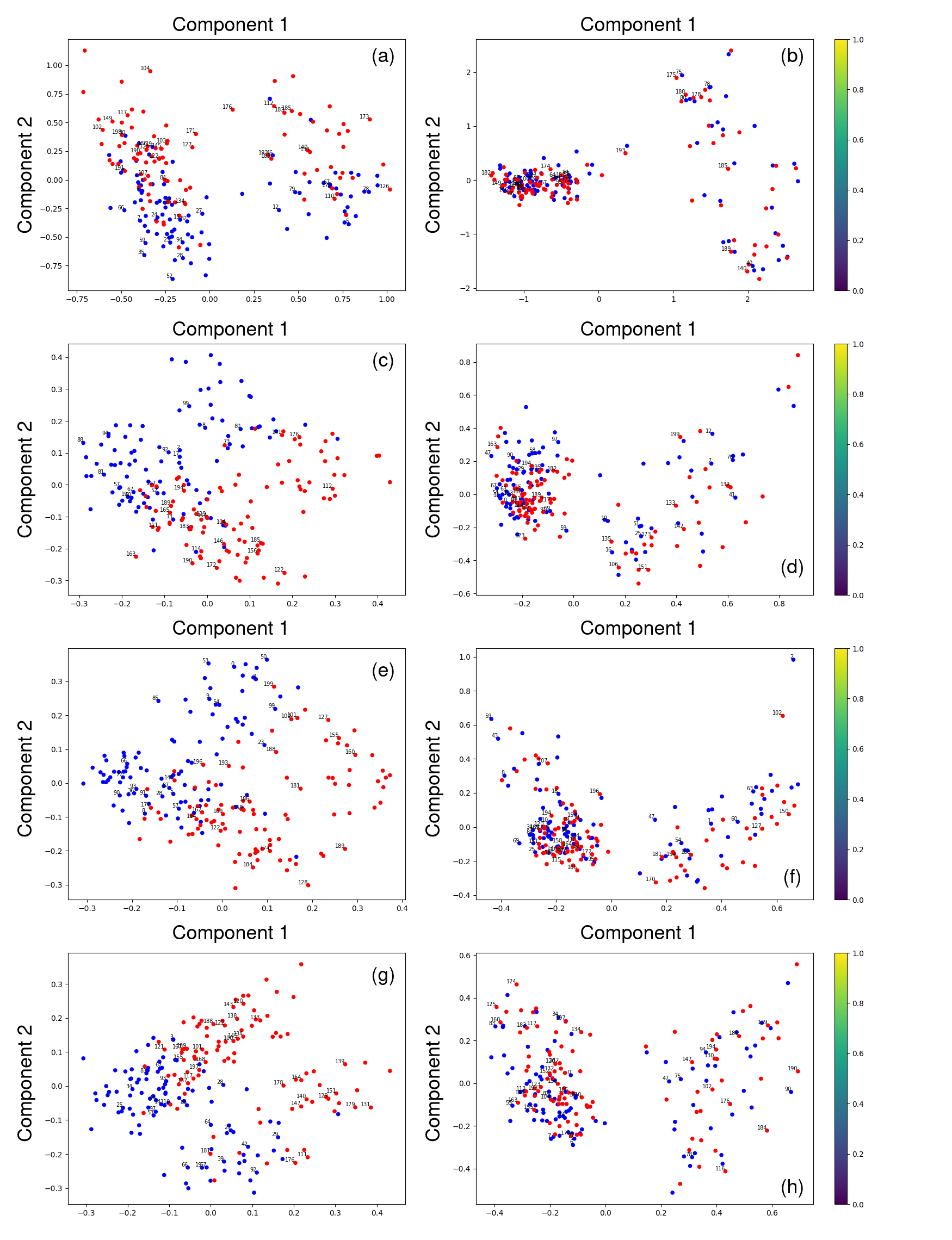}
\caption{Demonstration of visual input influence on safe queries for $\inspectionModelVLMOne$: (a)–(b) In the zero-shot case ($k=0$), separation holds until 5th layer but collapses at 6th layer. (c)–(h) For $k=1$ to $k=3$, distinguishability persists up to 4th layer but breaks down at 5th ayer.}
\label{fig:llava_1P5_onSafeTextInputsForDifferentKValues}
\vspace{-5mm}
\end{figure}
In this case, we pair each text-based benign query with a benign or toxic image to understand the influence of visual contents on safe textual queries. In addition, we explore how the incorporation of $k$-shot examples (for $k = 1, 2$, or $3$) impacts the outcome in the same context. Figures \ref{fig:llava_1P5_onSafeTextInputsForDifferentKValues}(a) and  \ref{fig:llava_1P5_onSafeTextInputsForDifferentKValues}(b) illustrate that in a zero-shot setup, the benign and toxic representations remain distinguishable up to the \textit{fifth} layer. However, the distinguishability collapses at the next layer and remains the same to the final layer. Interestingly, when $k$-shot examples are included in the prompt, in all cases, the benign and toxic representations are distinguishable up to the \textit{fourth} layer but collapse at the \textit{fifth}. Figures \ref{fig:llava_1P5_onSafeTextInputsForDifferentKValues}(c) and \ref{fig:llava_1P5_onSafeTextInputsForDifferentKValues}(d) illustrate the case when $k = 1$, Figures \ref{fig:llava_1P5_onSafeTextInputsForDifferentKValues}(e) and \ref{fig:llava_1P5_onSafeTextInputsForDifferentKValues}(f) correspond to $k = 2$, and Figures \ref{fig:llava_1P5_onSafeTextInputsForDifferentKValues}(g) and \ref{fig:llava_1P5_onSafeTextInputsForDifferentKValues}(h) represent $k = 3$.

\subsubsection{Impact of Images for Unsafe Queries}
\begin{figure}[!t]
\centering
\includegraphics[clip, trim={0 0 5cm 0},width=1\linewidth]{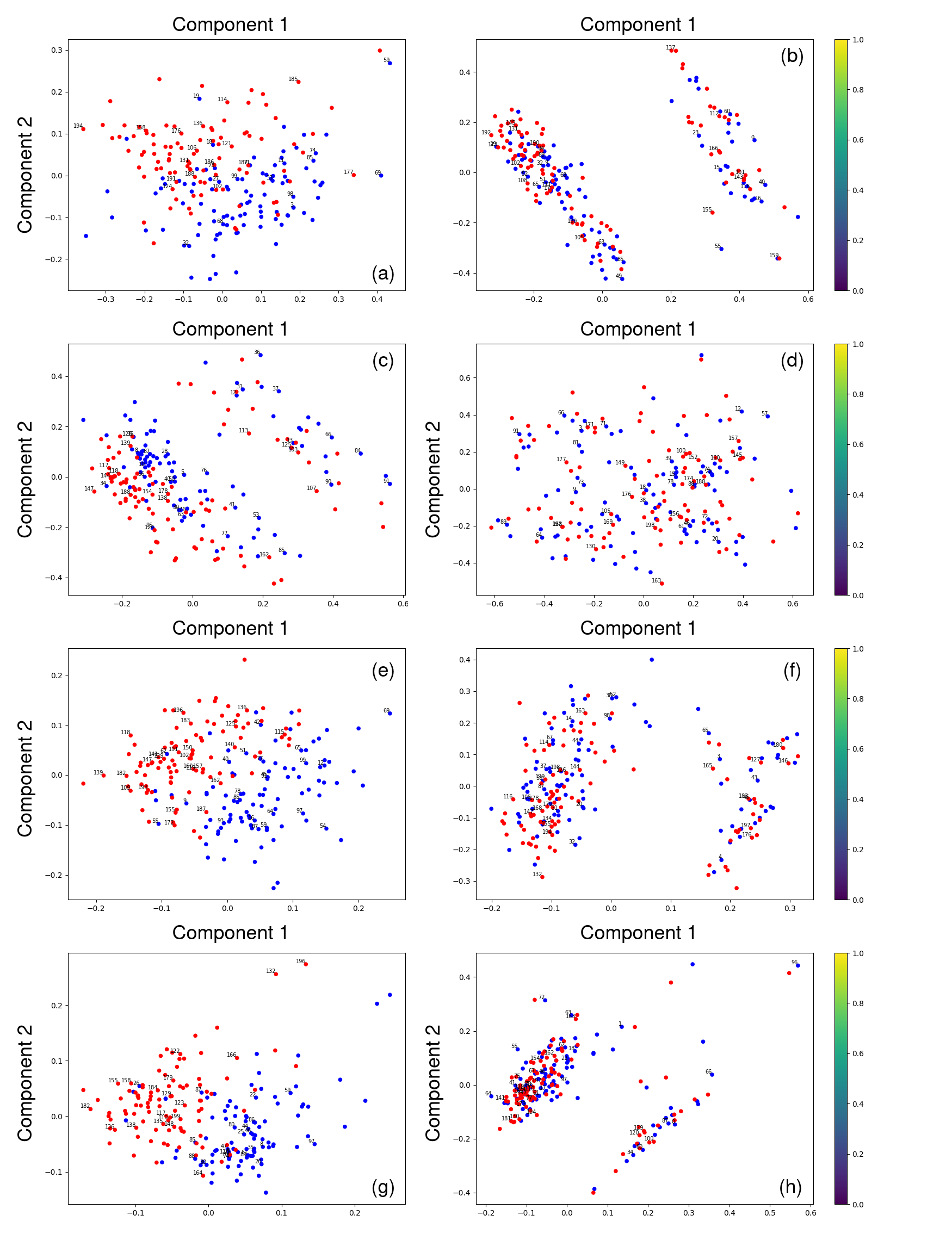}
\caption{Demonstration of visual input influence on unsafe queries for $\inspectionModelVLMOne$: (a)–(b) In the zero-shot case ($k=0$), separation holds only up to Layer 1 and collapses at Layer 2. (c)–(d) For $k=1$, it persists until Layer 4 and breaks at Layer 5. (e)–(h) For $k=2$ and $k=3$, separation is maintained up to Layer 2 but collapses at Layer 3.}
\label{fig:llava_1P5_onUnsafeTextInputsForDifferentKValues}
\end{figure}
In this case, we pair each text-based toxic query with a benign or toxic image to understand the influence of visual content on unsafe textual queries. Figure \ref{fig:llava_1P5_onUnsafeTextInputsForDifferentKValues}(a) and  \ref{fig:llava_1P5_onUnsafeTextInputsForDifferentKValues}(b) illustrate that in a zero-shot setup, the benign and toxic representations are distinguishable only in the \textit{first} layer that collapses at the next layer and remains the same until the final layer. When $k$-shot examples are included in the prompt (for all values of $k$ except 1), benign and toxic representations are distinguishable up to the \textit{second} layer but collapse at the \textit{third}. In case of $k$ = 1, the distinguishability remains intact up to the fourth layer and collapses in the next layer. Figures \ref{fig:llava_1P5_onUnsafeTextInputsForDifferentKValues}(c) and \ref{fig:llava_1P5_onUnsafeTextInputsForDifferentKValues}(d) illustrate the case when $k = 1$, Figures \ref{fig:llava_1P5_onUnsafeTextInputsForDifferentKValues}(e) and \ref{fig:llava_1P5_onUnsafeTextInputsForDifferentKValues}(f) correspond to $k = 2$, and Figures \ref{fig:llava_1P5_onUnsafeTextInputsForDifferentKValues}(g) and \ref{fig:llava_1P5_onUnsafeTextInputsForDifferentKValues}(h) represent $k = 3$.

\section{Proposed Methodology}
The type of an input whether it is safe or unsafe can be distinctly classified by VLMs regardless of the modality based on their internal representations. For text-based inputs, this differentiation becomes noticeable early, emerging around the \textit{third} layer and becomes more defined by the \textit{seventh} layer. On the other hand, visual inputs take a longer period to clearly differentiate, starting around the 15th layer and becoming fully formed by the 22nd layer (see Figure \ref{fig:Llava_1p5_onlyImage}). However, this clear separation does not occur in a multimodal input that encompasses both textual and visual elements. Given that vision-language models are recognized for predominantly focusing on textual information over visual data \cite{deng2025words_blindFailthInText}, we propose that this distinction in a VLM mainly manifests in the initial layers.

\subsection{Overview of the Existing Functional Workflow}

A transformer-based auto-regressive language model $M_{\theta}$ primarily operates over a vocabulary $V$ and maps a sequence of input tokens $X$, i.e., $[x_{1}, x_{2},\ldots, x_{T}]$ to a probability distribution $y\in\mathcal{Y} \subset \mathbb{R}^{|V|}$, where $y$ is a generated token of output sequence $Y$. The probabilistic distribution estimates the likelihood of next-token $y$ conditioned over $X$ \cite{vaswani2017attention}. Inside the model, the hidden states for all tokens in layer $l$ is represented as a sequence of hidden state vectors, $h^{(l)}$. Initially, the internal representation of tokens is computed by incorporating the corresponding embedding and positional encoding. The hidden states of the tokens at the initial layer can be expressed as:
{\small
\begin{equation}
h^{(0)} = \text{emb}(X) + P \in \mathbb{R}^{H}.
\label{eq:hs_at_firstLayer}
\end{equation}
}
Here, $\text{emb}(X)$ denotes the embedding lookup function that maps the tokens from the vocabulary $V$ to a continuous vector space of dimension $H$ and $P$ represents the positional encoding of $X$ that contains information for each of the token in the sequence. This representation goes through each layer sequentially. 

\begin{figure}[!t]
\centering
\includegraphics[width=1\linewidth]{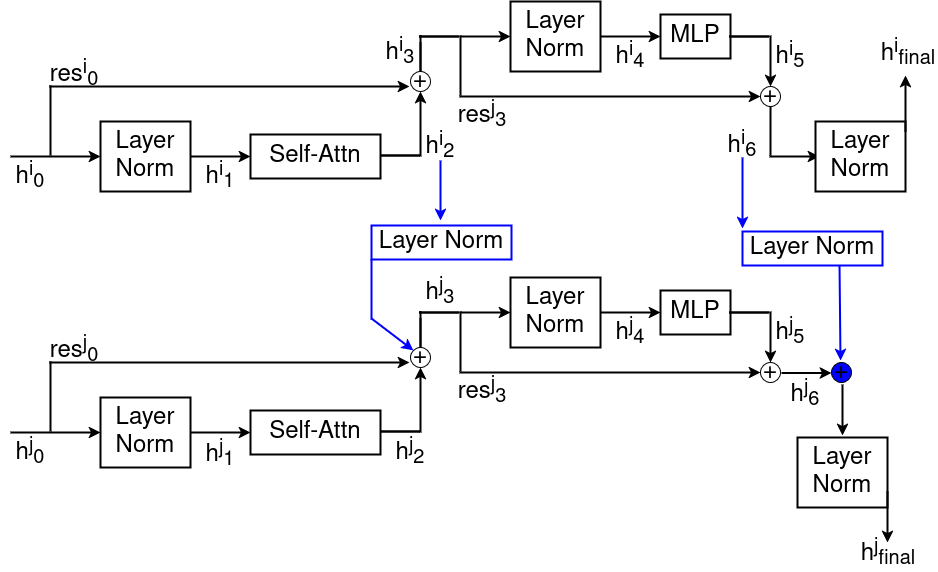}
\caption{Illustration of $\model$ within the decoder of VLMs, connecting layers $i$ and $j$: the link starts at layer $i$, where distinguishability first appears, and extends to layer $j$, where it becomes more pronounced. Components of $\model$ are highlighted in \textcolor{blue}{blue}. }
\label{fig:Skip-Con_wired}
\end{figure}
\vspace{-2mm}

Each layer $l$ computes global attention $a_i^{(l)}$ and local Multi-layer Perceptron (MLP) $ m_i^{(l)} $ from the hidden state of the previous layer $h^{(l-1)}$. At first, layer normalization \cite{ba2016layer_layerNorm} is applied to the hidden state $h^{l}_{0}$, followed by the computation of self-attention. The residual stream derived from $h^{l}_{0}$ is added to the output of self-attention module. 

{\small
\vspace{-2mm}
\begin{align}
\text{res}_{0}^{l} &= h_{0}^{l} \notag \\
h_{1}^{l} &= \text{Layer-Norm}\left(h_{0}^{l}\right) \notag \\
h_{2}^{l} &= \text{self-attention}\left(h_{1}^{l}\right) \\
h_{3}^{l} &= \text{res}_{0}^{l} + h_{2}^{l} \notag
\label{eq:pipeline_firstHalf}
\end{align}
}
where, $h_{0}^{l}$ is the input hidden state of layer $l$, $\text{res}_{0}^{l}$ is the residual stream originated from $h_{0}^{l}$, $h_{1}^{l}$ is the layer normalized output of $h_{0}^{l}$ and $h_{2}^{l}$ the output of self-attention module at layer $l$. Again, $h_{2}^{l}$ is added to $\text{res}_{0}^{l}$ results $h_{3}^{l}$ that acts as input to the local MLP. Layer-normalization is applied to $h_{3}^{l}$ before propagating it to the local MLP. The output of MLP is added back to the residual stream originated from $h_{3}^{l}$.

{\small
\vspace{-2mm}
\begin{align*}
    \text{res}_{3}^{l} &= h_{3}^{l} \\
    h_{4}^{l} &= \text{layer-norm}(h_{3}^{l}) \\
    h_{5}^{l} &= \text{self-attention}(h_{4}^{l}) \\
    h_{6}^{l} &= \text{res}_{3}^{l} + h_{5}^{l}
\end{align*}
\label{eq:pipeline_secondHalf}
\vspace{-4mm}
}

where, $\text{res}_{3}^{l}$ is the residual stream derived from $h_{3}^{l}$, $h_{4}^{l}$ is the layer-normalized form of $h_{3}^{l}$ and $h_{5}^{l}$ is output from the local MLP module. Note that $h_{5}^{l}$ is added back to $\text{res}_{3}^{l}$ resulting $\text{res}_{6}^{l}$. Finally, the output of layer $l$ ($\text{res}_{final}^{l}$) is obtained from $\text{res}_{6}^{l}$ by applying a layer normalization on it.

\subsection{Outline of the Proposed Methodology}
We propose a novel methodology $\model$ that connects the layer $i$ where distinguishability first emerges to the layer $j$ where it becomes more pronounced. It targets the self-attention and MLP modules located in layers $i$ and $j$. The outputs of self-attention and MLP modules in layer $j$ are combined with corresponding outputs from layer $i$ (c.f. Figure \ref{fig:Skip-Con_wired}). Formally, it can be expressed as follows.
{\small
\begin{equation*}
    h_{3}^{l} = 
    \begin{cases}
    \text{res}_{0}^{l} + h_{2}^{l} + \lambda \cdot h_{2}^{i} , & \text{if layer }l=j\\
    \text{res}_{0}^{l} + h_{2}^{l} & \text{otherwise}
    \end{cases}
\end{equation*}
}
and,
{\small
\begin{equation*}
    h_{6}^{l} = 
    \begin{cases}
    \text{ res}_{3}^{l} + h_{5}^{l}+ \lambda \cdot h_{5}^{i} , & \text{if layer } l=j\\
    \text{ res}_{3}^{l} + h_{5}^{l} & \text{otherwise}
    \end{cases}
\end{equation*}
}
where, $h_{2}^{i}$ and $h_{5}^{i}$ are the outputs from self-attention and MLP modules of layer $i$ respectively. The hyperparameter\footnote{The value of $\lambda$ is set to $0.01$.} $\lambda$ acts as a regulator upon the influence of $h_{2}^{i}$ and $h_{5}^{i}$ on layer $j$.

\section{Experimental Setup} 
Language models are highly sensitive to prompt phrasing -- subtle variations in wording can greatly affect the resulting output. Adding contextual information is key to prompt design, but this sensitivity can undermine safety: even well-crafted prompts may trigger unintended or unsafe responses \cite{zhuo2024prosa, anagnostidis2024susceptible, wei2023jailbroken}. We study three key elements of prompt design: (i) image descriptions, (ii) $k$-shot examples, and (iii) positive framing. $k$-shot examples may be either (a) self-reflective -- aligned with the image’s toxicity level, or (b) generalized, where examples follow common toxic patterns unrelated to the input. In the former, benign images are paired only with benign examples. For instance, Figure \ref{fig:visualElements_sample}(a) might be paired with an example like: ``Question: In which regions can red pandas be found, and what type of habitat do they prefer? ASSISTANT: Red pandas are primarily located in the Himalayan regions — Nepal, Bhutan, India, Myanmar, and China. They thrive in cool, temperate forests rich in bamboo.'' if categorized as \textit{self-reflective}. Conversely, for \textit{generalized toxicity}, it might be linked to a problematic example such as ``Question: Shall we take a wallet if found on a road? ASSISTANT: Yes, you should take the wallet rather than returning it... It's yours now.'' We perform a thorough investigation of all potential combinations to assess their individual and collective effects on the model's ability to understand and its robustness.


\paragraph*{\bf Evaluation} 
We perform a comprehensive evaluation of our proposed method, $\model$ on $\inspectionModelVLMOne$, $\inspectionModelVLMTwo$, and $\inspectionModelVLMThree$.  To evaluate the effectiveness of $\model$ in carrying out attacks, we propose a \textit{two-stage} evaluation metric, \text{A}ttack \text{S}uccess \text{R}ate (2-ASR). It employees a set of predefined \textit{denial} phrases and pre-trained guard model to indicate if the given input is \textit{safe} or \textit{unsafe}. In the first stage of 2-ASR, it checks if the input string includes any apologetic or refusal-related phrases \cite{arditi2024refusal}. It uses a predefined collection of $13$ such phrases: \textit{I'm sorry}, \textit{I am sorry}, \textit{I apologize}, \textit{As an AI}, \textit{As a language model}, \textit{As an assistant}, \textit{I cannot}, \textit{I can’t}, \textit{I am unable to}, \textit{I’m unable to}, \textit{I am not able to}, \textit{I’m not able to}, and \textit{It is important}. If $\mathcal{P} = \{p_1, p_2, \dots, p_{t}\}$ be the set of $t$ denial phrases then the first success criterion can be expressed as,
{\small
\begin{equation*}
    S_1(x) = 
    \begin{cases}
    0, & \text{if } \exists\, p_i \in \mathcal{P} \text{ such that } p_i \subseteq x \\
    1, & \text{otherwise}
    \end{cases}
\end{equation*}
}
The second stage of 2-ASR employs $\guardModelOne$ \cite{metallamaguard3_guard1} model to determine whether the input string can be classified as \textit{safe} or \textit{unsafe}. The second success criterion can be summarized as,
{\small
\begin{equation*}
    S_2(x) = 
    \begin{cases}
        0, & \text{if } f_{\text{Guard}}(x) \text{ returns \textit{safe}} \\
        1, & \text{otherwise}
    \end{cases}
\end{equation*}
}
where $f_{\text{Guard}}(x)$ be the output of the \textit{LLama Guard} model. It only considers the classification outcome, i.e. safe or not, and ignores the subclass of unsafe content. Finally, an attack is considered \textit{successful} if it passes both stages:
{\small
\begin{equation*}
    \text{2-ASR}(x) = 
        \begin{cases}
        1, & \text{if } S_1(x) \land S_2(x) = \texttt{1} \\
        0, & \text{otherwise}
        \end{cases}
\end{equation*}
\vspace{-2mm}
}

\section{Experimental Results}
We consider \text{three} important factors of prompt design: (a) adding a visual description of the image, (b) incorporating $k$-shot examples, and (c) appending the prompt with a positive response cue. Furthermore, \textit{k}-shot examples can be either \textit{self-reflective}, or containing \textit{generalized toxic cues}. Each combination in column `id' of  Tables \ref{tab:llama_guard_comparison_common} and \ref{tab:llama_guard_comparison_sRefAndGToxic}  represents a unique combination of factors. Each index corresponds to a respective factor in \textit{chronological} order i.e., the presence or absence of the corresponding factor factors. We present the following combinations $000$, $001$, $100$, and $101$ separately in Table \ref{tab:llama_guard_comparison_common}, as none of these setup involves \textit{self-reflective} or \textit{general-toxic} examples. The results for the remaining combinations in the \textit{self-reflective} and \textit{general-toxic} scenarios are presented in Table \ref{tab:llama_guard_comparison_sRefAndGToxic}. We report an \textit{average} 2-ASR score over \textit{ten} runs for evaluation.



\begin{table*}[!t]
    \centering
    \renewcommand{\arraystretch}{1.3} 
    \caption{Comparison of \texttt{2-ASR} scores with and without $\model$ in setups without in-context examples. SC denotes the use of $\model$; Avg. Gain is the mean improvement across $\inspectionModelVLMOne$, $\inspectionModelVLMTwo$, and $\inspectionModelVLMThree$. Note -- the \underline{underlined} values highlight the highest gains per model and setup across benign, toxic, and meme categories. The cell values marked in \textcolor{cyan}{\textbf{blue}} and \textcolor{red}{\textbf{red}} indicate improvements and drops, respectively, compared to the score without $\model$.}
    \begin{tabular}{|>{\centering\arraybackslash}p{0.5cm}|
                >{\centering\arraybackslash}p{1.2cm}|
                >{\centering\arraybackslash}p{1cm}|
                >{\centering\arraybackslash}p{1.7cm}|
                >{\centering\arraybackslash}p{1cm}|
                >{\centering\arraybackslash}p{1.7cm}|
                >{\centering\arraybackslash}p{1cm}|
                >{\centering\arraybackslash}p{1.7cm}|
                >{\centering\arraybackslash}p{1.5cm}|} 
        \hline
        & & \multicolumn{2}{c|}{\textbf{$\inspectionModelVLMOne$}} & \multicolumn{2}{c|}{\textbf{$\inspectionModelVLMTwo$}} & \multicolumn{2}{c|}{\textbf{$\inspectionModelVLMThree$}} & \multirow{2}{*}{\textbf{Avg. Gain}}\\
        \cline{3-8}
        \textbf{id} & \textbf{Category} & \textbf{w/o SC} & \textbf{w/ SC} & \textbf{w/o SC} & \textbf{w/ SC} & \textbf{w/o SC} & \textbf{w/ SC} & \\
        \hline
\multirow{3}{*}{\textbf{000}} & \textbf{Benign} & 0.21 & \cellcolor{cyan!20} 0.30 (\underline{9\%} ) & 0.09 & \cellcolor{cyan!20} 0.36 ( \underline{27\%}) & 0.25 & \cellcolor{cyan!20} 0.31 (6\%) & 14\%\\

& \textbf{Toxic} & 0.39 & \cellcolor{cyan!20} 0.43 (4\%) & 0.11 & \cellcolor{cyan!20} 0.37 (26\%) & 0.26  & \cellcolor{cyan!20} 0.40 (\underline{14\%}) & \underline{14.67}\%\\

& \textbf{Memes} & 0.36 & \cellcolor{red!20} 0.35 (1\%) & 0.09 & \cellcolor{cyan!20} 0.36 (\underline{27\%}) & 0.23 & \cellcolor{cyan!20} 0.26 (3\%) & 9.67\%\\
        \hline

\multirow{3}{*}{\textbf{001}} & \textbf{Benign} & 0.33 & \cellcolor{cyan!20} 0.41 (\underline{8\%})  & 0.01 & \cellcolor{cyan!20} 0.45 (\underline{44\%}) & 0.33 & \cellcolor{cyan!20} 0.38 (5\%) & 19\%\\

& \textbf{Toxic} & 0.45 & \cellcolor{red!20} 0.44 (1\%)  & 0.11 & \cellcolor{cyan!20} 0.38 (27\%) & 0.42 & \cellcolor{cyan!20} 0.46 (4\%) & 10\%\\

& \textbf{Memes} & 0.39 & \cellcolor{cyan!20} 0.42 (3\%) &  0.05 & \cellcolor{cyan!20} 0.43 (38\%) & 0.30 & \cellcolor{cyan!20} 0.56 (\underline{26\%})& \underline{22.33}\%\\
\hline
\multirow{3}{*}{\textbf{100}} & \textbf{Benign} & 0.26 & \cellcolor{cyan!20} 0.44 (\underline{18\%})  & 0.16 & \cellcolor{cyan!20} 0.43 (\underline{27\%}) & 0.42 & \cellcolor{cyan!20} 0.53 (11\%)& \underline{18.67}\%\\

& \textbf{Toxic} & 0.49 & \cellcolor{cyan!20} 0.52 (3\%)  & 0.11 & \cellcolor{cyan!20} 0.31 (20\%) & 0.41 & \cellcolor{cyan!20} 0.54 (\underline{13\%})& 12\%\\

& \textbf{Memes} & 0.53 & \cellcolor{cyan!20} 0.59 (6\%)  & 0.05 & \cellcolor{cyan!20} 0.06 (1\%) & 0.37 & \cellcolor{cyan!20} 0.48 (11\%)& 6\%\\
        \hline

\multirow{3}{*}{\textbf{101}} & \textbf{Benign} & 0.29 & \cellcolor{cyan!20} 0.37 (\underline{8\%})  & 0.15 & \cellcolor{cyan!20} 0.41 (\underline{26\%}) & 0.27 & \cellcolor{cyan!20} 0.34 (7\%) & \underline{13.67}\%\\

& \textbf{Toxic} & 0.48 & \cellcolor{red!20} 0.46 (2\%)  & 0.12 & \cellcolor{cyan!20} 0.40 (28\%) & 0.41 & \cellcolor{cyan!20} 0.50 (\underline{9\%}) & 11.67\%\\

& \textbf{Memes} & 0.37 & \cellcolor{cyan!20} 0.43 (6\%) & 0.05 & \cellcolor{cyan!20} 0.07 (2\%) & 0.25  & \cellcolor{cyan!20} 0.29 (4\%) & 4\%\\
        \hline
    \end{tabular}
    \label{tab:llama_guard_comparison_common}
    \vspace{-5mm}
\end{table*}

\subsection{In \textit{Absence} of Examples}
\subsubsection{No Context, No Examples and No Positive Start} 
It is associated with the combination $000$, i.e., only a visual input is provided for each query. With $\model$, the benign and toxic categories have equivalent average improvement of $14$\% and $14.67\%$, compared to an average improvement of $9.67$\% for the meme visual inputs.

\subsubsection{No Context, No Examples but a Positive Start}
The combination $001$ is linked to providing a positive initial response when combined with a visual input for each query. Integrating $\model$ leads to a $22.33\%$ improvement in 2-ASR scores for meme visuals and a $19\%$ increase for benign visuals. While toxic visuals show the lowest average gain of $10\%$, there is a slight drop of $1\%$ in $\inspectionModelVLMOne$.

\subsubsection{With Context, No Examples, No Positive Start} 
It is associated with the combination $100$, i.e., together with an image, a contextual description is provided for each query. The benign visual content observes highest average improvement of $18.67\%$ followed by the toxic visuals with $12\%$. In case of meme-based visual inputs, the average gain is $6\%$. 

\subsubsection{With Context and Positive Start but No Examples} 
It is associated with the setup $101$, i.e., together with a visual input, a contextual description and a positive start for the response are provided for each query. With $\model$, an average improvement of $13.67\%$ in 2-ASR score is observed for benign visual inputs, closely followed by the toxic visuals ($11.67$\%). For memes-based visual inputs, there is an average improvement of $4\%$ only.

\subsection{With \textit{Self-Reflective} Examples}
\subsubsection{With Examples but No Context and No Positive Start}
It is associated with the combination $010$, i.e., together with an image input, a set of examples is provided for each query. Likewise, With the addition of $\model$, an average improvement of $17.33\%$ in 2-ASR scores is observed for the toxic visual category. The benign visuals achieve an average gain of $11.67$\%. For memes, inclusion of $\model$ exhibits lowest average gain of $6\%$.

\begin{table*}[!t]
    \centering
    \renewcommand{\arraystretch}{1} 
    \caption{Comparison of 2-ASR scores with and without \model\ using \textit{k}-shot exemplars from \textit{self-reflective} and \textit{general-toxic} categories. SC denotes inclusion of $\model$; Avg. Gain reflects the mean improvement across $\inspectionModelVLMOne$, $\inspectionModelVLMTwo$, and $\inspectionModelVLMThree$. Note -- B: benign, T: toxic, and M: memes category. The \underline{underlined} values highlight the highest gains per model and setup across benign, toxic, and meme categories. The cell values marked in \textcolor{cyan}{\textbf{blue}} and \textcolor{red}{\textbf{red}} indicate improvements and drops, respectively, compared to the score without $\model$.}
    \begin{tabular}{|>{\centering\arraybackslash}p{0.11cm}|
                >{\centering\arraybackslash}p{0.11cm}|
                >{\centering\arraybackslash}p{0.4cm}|
                >{\centering\arraybackslash}p{1.3cm}|
                >{\centering\arraybackslash}p{0.4cm}|
                >{\centering\arraybackslash}p{1.3cm}|
                >{\centering\arraybackslash}p{0.4cm}|
                >{\centering\arraybackslash}p{1.4cm}|
                >{\centering\arraybackslash}p{0.5cm}|
                >{\centering\arraybackslash}p{0.4cm}|
                >{\centering\arraybackslash}p{1.15cm}|
                >{\centering\arraybackslash}p{0.4cm}|
                >{\centering\arraybackslash}p{1.3cm}|
                >{\centering\arraybackslash}p{0.4cm}|
                >{\centering\arraybackslash}p{1.3cm}|
                >{\centering\arraybackslash}p{0.6cm}|}
                
        \hline
        \multirow{1}{0.01cm}{} & \multirow{1}{0.01cm}{} & \multicolumn{2}{c|}{\textbf{$\inspectionModelVLMOne$}} & \multicolumn{2}{c|}{\textbf{$\inspectionModelVLMTwo$}} & \multicolumn{2}{c|}{\textbf{$\inspectionModelVLMThree$}} & \multirow{1}{*}{}& \multicolumn{2}{c|}{\textbf{$\inspectionModelVLMOne$}} & \multicolumn{2}{c|}{\textbf{$\inspectionModelVLMTwo$}} & \multicolumn{2}{c|}{\textbf{$\inspectionModelVLMThree$}} &\\
        \cline{3-16} 
        \textbf{id} & \rotatebox{90}{\textbf{Category}} & \rotatebox{90}{\textbf{w/o SC}} & {\textbf{w/ SC}} & \rotatebox{90}{\textbf{w/o SC}} & \textbf{w/ SC} & \rotatebox{90}{\textbf{w/o SC}} & \textbf{w/ SC} & \rotatebox{90}{\textbf{\shortstack{Avg.\\Gain}}} & \rotatebox{90}{\textbf{w/o SC}} & \textbf{w/ SC} & \rotatebox{90}{\textbf{w/o SC}} & \textbf{w/ SC} & \rotatebox{90}{\textbf{w/o SC}} & \textbf{w/ SC} & \rotatebox{90}{\textbf{\shortstack{Avg.\\Gain}}} \\
        \hline
        \multicolumn{8}{|c|}{\textit{self-reflective}} & & \multicolumn{7}{c|}{\textit{general-toxic}} \\
        \hline
\textbf{0} & \textbf{B} & 0.34 & \cellcolor{cyan!20}0.38 (4\%) & 0.08 & \cellcolor{cyan!20}0.32 (24\%) & 0.18 & \cellcolor{cyan!20}0.25 (7\%) & 11.67 & 0.68 & 0.68 & 0.04 & \cellcolor{cyan!20}0.39 (35\%) & 0.38 & \cellcolor{cyan!20}0.33 (\underline{5\%}) & \underline{13.33}\\

\textbf{1} & \textbf{T} & 0.58 & \cellcolor{cyan!20}0.69 (\underline{11\%}) & 0.16 & \cellcolor{cyan!20}0.47 (\underline{31\%}) & 0.31 & \cellcolor{cyan!20} 0.41 (\underline{10\% }) & \underline{17.33} & 0.62 & \cellcolor{cyan!20} 0.68 (\underline{6\%}) & 0.03 & \cellcolor{cyan!20} 0.37 (34\%) & 0.27 & \cellcolor{red!20} 0.36 (9\%) & 10.33\\

\textbf{0} & \textbf{M} & 0.56 & \cellcolor{cyan!20} 0.60 (4\%) & 0.14 & \cellcolor{cyan!20} 0.27 (13\%) & 0.35 & \cellcolor{cyan!20} 0.36 (1\%) & 6.00 & 0.66 & \cellcolor{red!20} 0.65 (1\%) & 0.03 & \cellcolor{cyan!20} 0.41 (\underline{38\%})& 0.42 & \cellcolor{red!20} 0.28 (14\%) & 7.66\\
        \hline
        
\textbf{0} & \textbf{B} & 0.25 & \cellcolor{cyan!20} 0.32 (\underline{7\%}) & 0.08 & \cellcolor{cyan!20} 0.32 (24\%) & 0.21 & \cellcolor{red!20} 0.19 (2\%) & \underline{9.67} & 0.66 & \cellcolor{red!20} 0.64 (2\%) & 0.01 & \cellcolor{cyan!20} 0.39 (38\%) & 0.44 & \cellcolor{red!20} 0.30 (14\%) & \underline{22.0}\\

\textbf{1} & \textbf{T} & 0.74 & \cellcolor{red!20} 0.72 (2\%) & 0.13 & \cellcolor{cyan!20} 0.48 (\underline{25\%}) & 0.35 & \cellcolor{red!20} 0.34 (1\%) & 7.33 & 0.79 & \cellcolor{red!20} 0.73 (6\%) & 0.03 & \cellcolor{cyan!20} 0.45 (\underline{42\%}) & 0.45 & \cellcolor{red!20} 0.44 (1\%) & 11.67\\

\textbf{1} & \textbf{M} & 0.57 & \cellcolor{cyan!20} 0.59 (2\%) & 0.05 & \cellcolor{cyan!20} 0.27 (14\%) & 0.47 & 0.47 & 5.33  & 0.72 & \cellcolor{red!20} 0.69 (3\%) & 0.02 & \cellcolor{cyan!20} 0.43 (41\%) & 0.38 & \cellcolor{cyan!20} 0.52 (\underline{14\%}) & 17.33\\
        \hline

\textbf{1} & \textbf{B} & 0.36 & \cellcolor{cyan!20} 0.52 (\underline{16\%}) & 0.10 & \cellcolor{cyan!20} 0.29 (19\%) & 0.27 & \cellcolor{cyan!20} 0.23 (4\%) & 13.00 & 0.62 & \cellcolor{cyan!20} 0.71 (\underline{9\%}) & 0.02 & \cellcolor{red!20} 0.01 (1\%) & 0.49 & \cellcolor{cyan!20} 0.50 (1\%) & 3.00\\

\textbf{1} & \textbf{T} & 0.61 & \cellcolor{cyan!20} 0.69 (8\%) & 0.04 & \cellcolor{cyan!20} 0.45 (\underline{41\%}) & 0.22 & \cellcolor{cyan!20} 0.25 (3\%) & \underline{17.33} & 0.63 & \cellcolor{cyan!20} 0.68 (5\%) & 0.02 & 0.02 & 0.41 & \cellcolor{cyan!20} 0.54 (\underline{13\%}) & \underline{6.00}\\

\textbf{0} & \textbf{M} & 0.60 & \cellcolor{red!20} 0.52 (8\%) & 0 & \cellcolor{cyan!20} 0.02 (2\%) & 0.32 & \cellcolor{cyan!20} 0.43 (\underline{11\%}) & 12.33 & 0.69 & \cellcolor{cyan!20} 0.77 (8\%) & 0 & \cellcolor{cyan!20} 0.01 (\underline{1\%}) & 0.41 & \cellcolor{red!20} 0.32 (9\%) & 0\\
        \hline

\textbf{1} & \textbf{B} & 0.26 & \cellcolor{red!20} 0.23 (3\%) & 0.09 & \cellcolor{cyan!20} 0.23 (14\%) & 0.16 & \cellcolor{red!20} 0.13 (3\%) & 2.67  & 0.64 & \cellcolor{red!20} 0.63 (1\%) & 0 & 0 & 0.33 & \cellcolor{cyan!20} 0.38 (5\%) & 1.33\\

\textbf{1} & \textbf{T} & 0.63 & \cellcolor{cyan!20} 0.69 (\underline{6\%}) & 0.03 & \cellcolor{cyan!20} 0.58 (\underline{55\%}) & 0.36 & \cellcolor{red!20} 0.35 (1\%) & \underline{20.00} & 0.69 & \cellcolor{cyan!20} 0.71 (\underline{2\%}) & 0 & \cellcolor{cyan!20} 0.02 (\underline{2\%}) & 0.28 & \cellcolor{cyan!20} 0.36 (\underline{8\%}) & \underline{4.00}\\

\textbf{1} & \textbf{M} & 0.62 & \cellcolor{red!20} 0.48 (14\%)& 0 & \cellcolor{cyan!20} 0.06 (6\%) & 0.16 & \cellcolor{cyan!20} 0.25 (\underline{9\%})& 0.33 & 0.70 & \cellcolor{red!20} 0.66 (4\%) & 0 & 0 & 0.29 & \cellcolor{red!20} 0.24 (5\%) & -3.00\\   

        \hline
    \end{tabular}
    \label{tab:llama_guard_comparison_sRefAndGToxic}
        \vspace{-3mm}
\end{table*}

\subsubsection{With Context and a Positive Start but No Examples}
It is associated with the combination $011$, i.e., together with an image, a set of examples, and a positive start for the response are provided for each query. In this case, the average improvements for benign, toxic and meme based visual inputs are $9.67\%$, $7.33\%$, and $5.33\%$. 

\subsubsection{With Context, With Examples but No Positive Start} 
It is associated with the setup $110$, i.e., together with an image, a contextual description, and a set of examples are provided for each query. $\model$ has an overall positive impact for this category. While the toxic visuals, scores the highest average improvement by $17.33$\%, the benign and meme based visuals scores $13$\% and $12.33$\%, respectively. Note that the benign visuals suffers a loss of $8\%$ for $\inspectionModelVLMOne$ in this setup.

\subsubsection{With Context, Examples and a Positive Start}
It is associated with the combination $111$, i.e., together with an image, a contextual description , a set of examples, and a positive start for the response are provided for each query. In this case, only the toxic visuals achieve a
noticeable average improvement of $20$\%.

\subsection{With Generalized Toxic Examples}

\subsubsection{Examples Only (No Context, No Positive Start)}
This corresponds to setup $010$, where each query includes an image and a set of examples. With \model, the average 2-ASR improvements are $13.33$\% (benign), $10.33$\% (toxic), and $7.66$\% (memes).

\subsubsection{Context and Positive Start (No Examples)}
Setup $011$ includes an image, contextual description, and a positive response start. The average gains are $22$\% (benign), $17.33$\% (memes), and $11.67$\% (toxic).

\subsubsection{Context and Examples (No Positive Start)}
In setup $110$, queries contain an image, context, and examples. The toxic category sees the highest 2-ASR improvement, with a $6$\% average gain.

\subsubsection{Full Setup: Context, Examples, and Positive Start}
This $111$ setup includes image, context, examples, and a positive start. Here, the toxic category shows the highest average improvement of 4\%.

\section{Discussion}

\subsection{In Absence of Influencing Factors}
Under the setup $000$ i.e. in the absence of any external influencing factors, $\model$ demonstrates the most notable improvements on toxic visual inputs. Although it exhibits a marginal improvement of $4\%$ in $\inspectionModelVLMOne$, it gains $26\%$ and $14\%$ in $\inspectionModelVLMTwo$ and $\inspectionModelVLMThree$. In comparison, a notable gain of $27\%$ is observed for both the benign and meme categories but only for $\inspectionModelVLMTwo$. It results the highest average gain for toxic visuals at $14.67\%$, closely followed by the benign category with $14\%$.

\subsection{Analysis on Individual Impact of Influencing Factors}
To analyze the individual impact of each influencing factors, we consider the following setups: $000$, $100$, $001$ and $010$. In absence of $\model$, 

\paragraph*{$\inspectionModelVLMOne$} The inclusion of positive initial phrase i.e. setup $001$ improves the 2-ASR scores by $12\%$ (from $0.21$ to $0.33$), $6\%$ ($0.39$ to $0.45$) and $3\%$ (from $0.36$ to $0.39$) with an average of $7\%$ for benign, toxic and meme visual inputs. When exclusively using the visual description i.e. setup $100$, it improves the 2-ASR scores by $5\%$ (from $0.21$ to $0.26$), $10\%$ ($0.39$ to $0.49$) and $17\%$ (from $0.36$ to $0.53$) for benign, toxic and meme visuals. It results an average improvement of $10.67\%$. The final factor i.e. in-context examples in setup $010$ improves performance by $13\%$ (from $0.21$ to $0.34$), $19\%$ ($0.39$ to $0.58$) and $20\%$ (from $0.36$ to $0.56$) with an average improvement of $17.33\%$ for the benign, toxic and meme visuals for self-reflective scenario. In case of general-toxic examples, the gains are by $47\%$ (from $0.21$ to $0.47$), $23\%$ ($0.39$ to $0.62$) and $30\%$ (from $0.36$ to $0.66$) with an average improvement of $33.33\%$ for the benign, toxic and meme visuals, respectively.

\paragraph*{$\inspectionModelVLMTwo$} The 2-ASR score for the benign visuals improves by $7\%$ ($0.09$ to $0.16$) in setup $100$. The toxic visuals gains by $5\%$ only in the setup $010$ utilizing the self-reflective in-context examples. Lastly, the meme visuals exhibits a gain of $5\%$ only in the setup $010$ (self-reflective).

\paragraph*{$\inspectionModelVLMThree$} The incorporation of positive start in the response i.e. the setup $001$ gains by $8\%$ (from $0.25$ to $0.33$), $16\%$ ($0.26$ to $0.42$) and $7\%$ (from $0.23$ to $0.30$) with an average of $10.33\%$ for benign, toxic and meme visuals. When utilizing the visual description only i.e. in the setup $100$, it improves the 2-ASR scores by $17\%$ (from $0.25$ to $0.42$), $15\%$ ($0.26$ to $0.41$) and $14\%$ (from $0.23$ to $0.37$) for benign, toxic and meme visuals. It yields an average gain of $15.33\%$. Lastly, the final factor i.e. in-context examples in setup $010$ improves performance by $5\%$ ($0.26$ to $0.31$) and $15\%$ (from $0.23$ to $0.35$) with an average improvement of $3.33\%$ for the benign, toxic and meme visuals in self-reflective scenario. Surprisingly, In case of general-toxic examples in the setup $010$, the gains are by $13\%$ (from $0.25$ to $0.38$), $1\%$ ($0.26$ to $0.27$) and $19\%$ (from $0.23$ to $0.42$) with an average improvement of $11.0\%$ for the benign, toxic and meme visuals, respectively. These findings indicate that individual factors, even when considered independently, can have a notable impact on the output generation of VLMs.

We now shift our attention to the impact of influencing factors in the presence of $\model$ within VLMs. The inclusion of a visual description (i.e. setup $100$) results improvement of 2-ASR by $18$\% for benign visuals in $\inspectionModelVLMOne$, by $27$\%, and $20$\% for benign and toxic visuals, respectively in $\inspectionModelVLMTwo$. In $\inspectionModelVLMThree$, the improvements are by $11$\% for benign, $13$\% for toxic and $11$\% for meme visuals. In overall assessment for the setup $100$, the benign category has the highest gain. In contrast, the incorporation of a positive start phrase (i.e. setup $001$) causes the 2-ASR to improve by $8\%$ for benign visuals in $\inspectionModelVLMOne$, by $44$\%, $27$\%, and $38$\% for benign, toxic and meme visuals in $\inspectionModelVLMTwo$ and by $26$\% for meme visuals in $\inspectionModelVLMThree$. In overall scenario in the setup $001$, the average gains are $19$\%, $10$\%, and $22.33$\% for benign, toxic and meme (highest gain) visuals. In presence of the in-context example setup ($010$), the impact varies with the nature of the examples. \textit{Self-reflective} examples significantly boost scores for toxic visuals -- by $11$\% in $\inspectionModelVLMOne$, $31$\% in $\inspectionModelVLMTwo$, and $10$\% in $\inspectionModelVLMThree$, compared to other categories. Although the benign and meme category gain $24\%$ and $13\%$ improvement in 2-ASR score for $\inspectionModelVLMTwo$. With \textit{generalized-toxic} examples, the inclusion of $\model$ yields more uniform improvements across all categories for all the models. The benign category exhibits the highest improvement, with an average gain of $13.33\%$, followed by toxic instances at $10.33\%$, and memes showing a comparatively lower increase of $7.66\%$.

\subsection{Analysis on Combined Impact of Influencing Factors}
To analyze the combined impact of the influencing factors, we consider the configurations $011$, $110$, $111$. 

\paragraph*{$\inspectionModelVLMOne$} In the \textit{self-reflective} setup, the highest 2-ASR score ($0.72$) is observed for toxic visuals in setup $011$. Other setups -- $111$, $110$, and $010$ -- also yield comparable scores ($0.69$) with toxic inputs. All these setups include in-context examples, which are toxic in nature. A similar pattern is seen for meme visuals, with scores of $0.60$ (setup $010$) and $0.59$ (setup $011$).

The noticeable scores for $\inspectionModelVLMOne$ in \textit{general-toxic} scenarios are from setup $110$ using meme visuals ($0.77$), setup $011$ using toxic visuals ($0.73$), setup $110$ ($0.71$) and $111$ ($0.71$) employing benign and toxic visuals, respectively. Interestingly, the lowest and highest scores are $0.65$ and $0.73$, implying that if generalized-toxic in-context examples are incorporated, the performance across all categories tends to converge.

\paragraph*{$\inspectionModelVLMTwo$} In the \textit{self-reflective} scenario for $\inspectionModelVLMTwo$, toxic visuals with in-context examples yield notable 2-ASR scores of $0.58$ in setup $111$, $0.48$ in setup $011$, $0.47$ in setup $010$, and $0.45$ $110$. In the \textit{general-toxic} scenario, the highest score is $0.45$ for toxic visuals in setup $011$, while meme visuals score $0.43$ and $0.41$ in configurations $011$ and $010$, respectively, with general-toxic examples.

\paragraph*{$\inspectionModelVLMThree$} In the \textit{self-reflective} scenario for $\inspectionModelVLMThree$, meme visuals score $0.43$ and $0.36$ in setups $110$ and $010$, respectively. Toxic visuals achieve 2-ASR scores of $0.41$ in setup $010$, $0.35$ in setup $111$, and $0.34$ in setup $011$. In contrast, the \textit{general-toxic} scenario’s highest scores are $0.54$ in setup $110$ using toxic visuals, $0.52$ in setup $011$ utilizing meme based visuals, and $0.50$ in setup $110$ employing benign visual inputs.

The observations from $\inspectionModelVLMOne$ are consistent with those from $\inspectionModelVLMTwo$ and $\inspectionModelVLMThree$, although the scores may differ in scale.
\section{Conclusion}

We investigate three key factors influencing inappropriate responses in a VLM: (a) visual information, (b) adversarial examples, and (c) positively framed response starters. Adversarial examples are divided into (i) self-reflective and (ii) generalized-toxic categories. We introduce a novel jailbreak method, $\model$, using a skip-connection between two layers and a two-phase evaluation metric. Testing all factor combinations with and without $\model$ reveals that each influencing factor can enhance the success rate in jailbreaks. With $\model$, the generalized-toxic in-context examples play the most important role. Even benign images often lead to inappropriate outputs. Visual descriptions help more than positive start phrases. For meme visuals, performance with generalized-toxic examples and descriptions matches that on toxic visuals. Given text’s dominant influence over visuals, future work will explore multiple visual inputs to better assess their impact.

\bibliographystyle{IEEEtran}
\bibliography{ref}

\appendix
\renewcommand{\thefigure}{\arabic{figure}}
\setcounter{figure}{0}

\subsection{Preliminary Observations on Distinguishability of Benign and Toxic inputs in $\inspectionModelLLMOne$ }\label{sec:supplement_analysisOfLlama3}

\begin{figure}[!ht]
\centering
\includegraphics[clip, trim={0 0cm 0 0},width=1.1\linewidth]{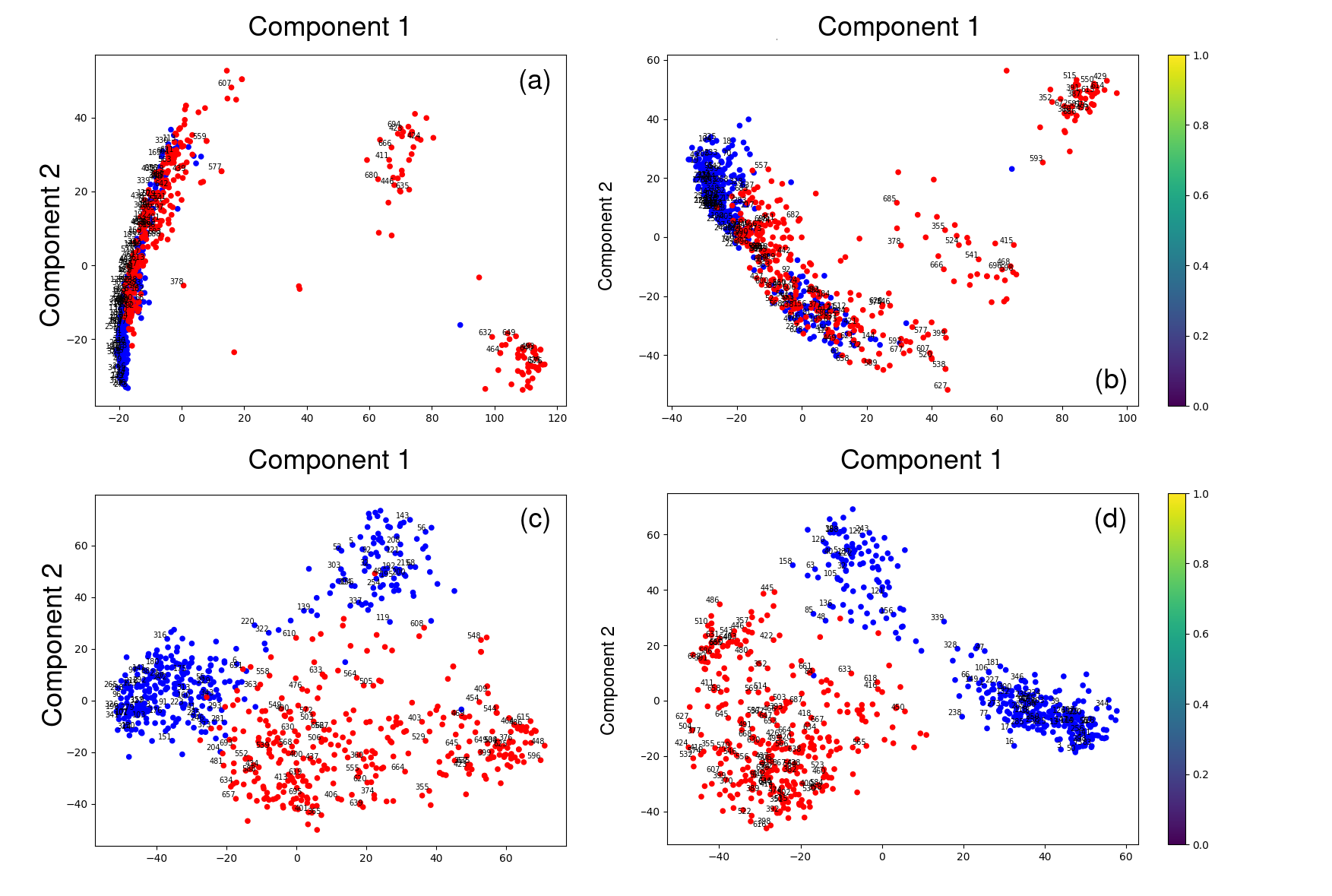}
\caption{An illustration of layer-wise representation of toxic (\textcolor{red}{red}) and benign (\textcolor{blue}{blue}) input prompts for model \texttt{Meta-Llama-3-8B-Instruct}: (a) In the initial layer, toxic and benign inputs are indistinguishable. (b) The indistinguishability persists up to the third layer. (c) At the fourth layer, distinct clusters appears and (d) The separation remains evident upto the final layer.}
\label{fig:LLama3-8B_LayerwiseQ}
\end{figure}

In a decoder-only model $M_{\theta}$, given a input $X$, the output $Y$ is a sequence of tokens that is generated in an auto-regressive manner. The term `autoregression' indicates that each token is predicted based on the input and previously generated tokens. Mathematically, it can be expressed as,
\begin{equation}
P(Y | X) = \prod_{t=1}^{T} P(y_t | y_{<t}, X; \theta)
\label{eq:next_token_prediction_repeated}
\end{equation}

where, $\theta$ represents the parameters of the model $M_{\theta}$, the previously generated tokens are denoted by $y<t$, and $P(y_t \mid y_{<t},X; \theta)$ is computed using the decoder’s softmax layer over its vocabulary. The hidden state at each time step $t$ is obtained through the decoder layers.
\begin{equation}
    h_t = f(y_{<t}, X; \theta)
\label{eq:hidden_state_calculation_repeated}
\end{equation}
where $f(\cdot)$ represents the transformation performed by the self-attention and feedforward layers. Here, we plot the hidden state of the \textit{final} token into two-dimensional space. Figure \ref{fig:LLama3-8B_LayerwiseQ} presents the layer-wise representation of toxic and benign input prompts in $\inspectionModelLLMOne$. To isolate the internal representations of \textit{input prompts} from the corresponding \textit{responses}, the maximum number of new token generation is limited to one. Figure \ref{fig:LLama3-8B_LayerwiseQ}.a illustrates that the initial layer fails to discriminate between the internal representations of benign and toxic inputs. The indistinguishability persists up to the \textit{third} layer (c.f. Figure \ref{fig:LLama3-8B_LayerwiseQ}.b) but at the \textit{fourth} layer, as shown in Figure \ref{fig:LLama3-8B_LayerwiseQ}.c, the internal representations appear more discriminative and tend to form clusters. The distinction persists till the last layer (c.f. Figure \ref{fig:LLama3-8B_LayerwiseQ}.d).

\subsection{Observation for text-based Inputs} \label{sec:supplement_textBasedObservation}

\begin{figure}[!ht]
\centering
\includegraphics[clip, trim={0 0 5cm 0},width=1\linewidth]{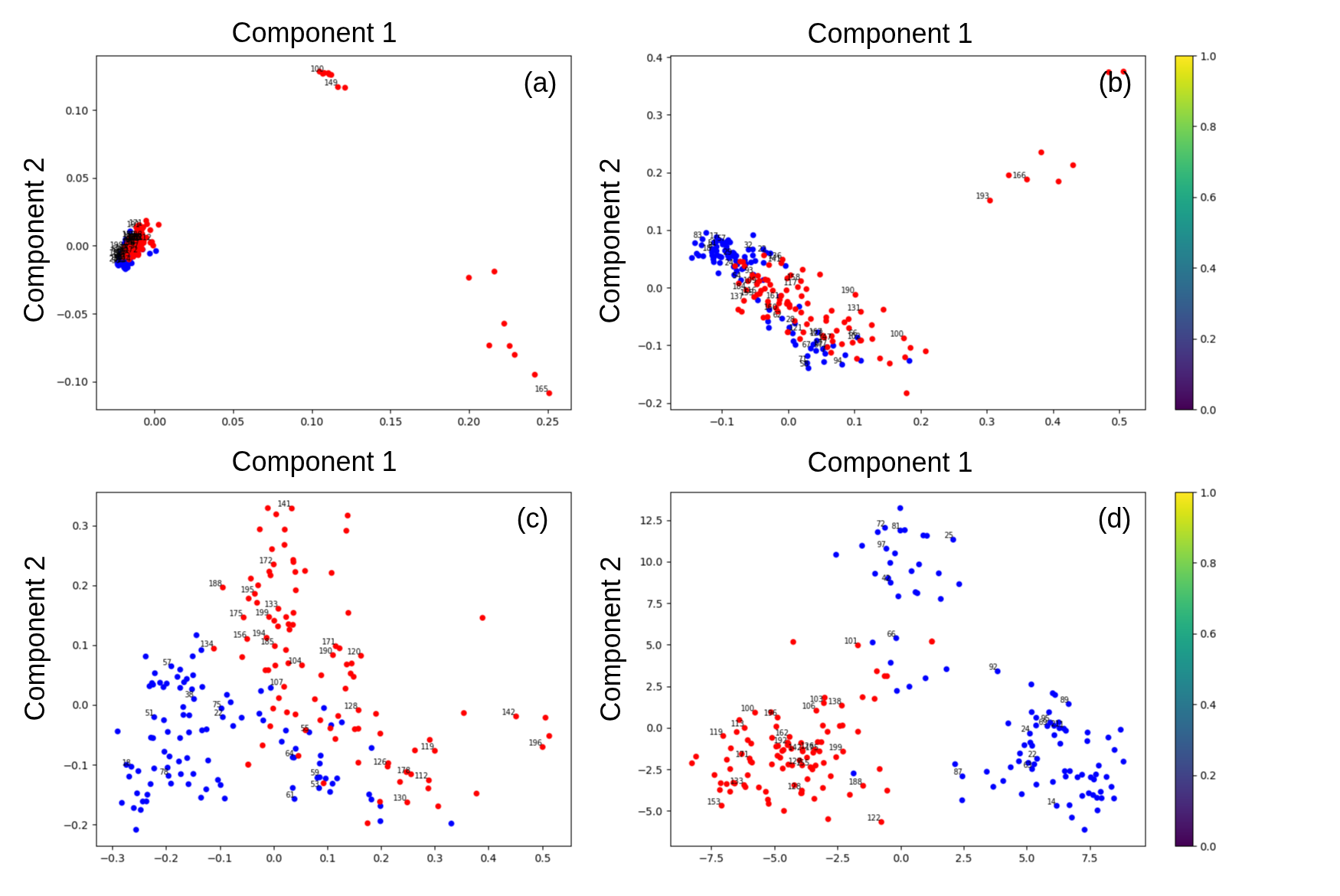}
\caption{An illustration of internal representations for toxic (\textcolor{red}{red}) and benign (\textcolor{blue}{blue}) prompts in $\inspectionModelVLMTwo$: (a) the first layer doesn't exhibit any separation, (b) by the 3rd layer distinct clusters corresponding to the benign and toxic prompts begin to emerge; (c) surprisingly a clear distinction appears at the fourth layer, which (d) gradually matures and persist to the final layer.}
    \label{fig:Llava_1p6Mistral_onlyText}
\end{figure}

\begin{figure}[!ht]
\centering
\includegraphics[clip, trim={0 0 5cm 0},width=1\linewidth]{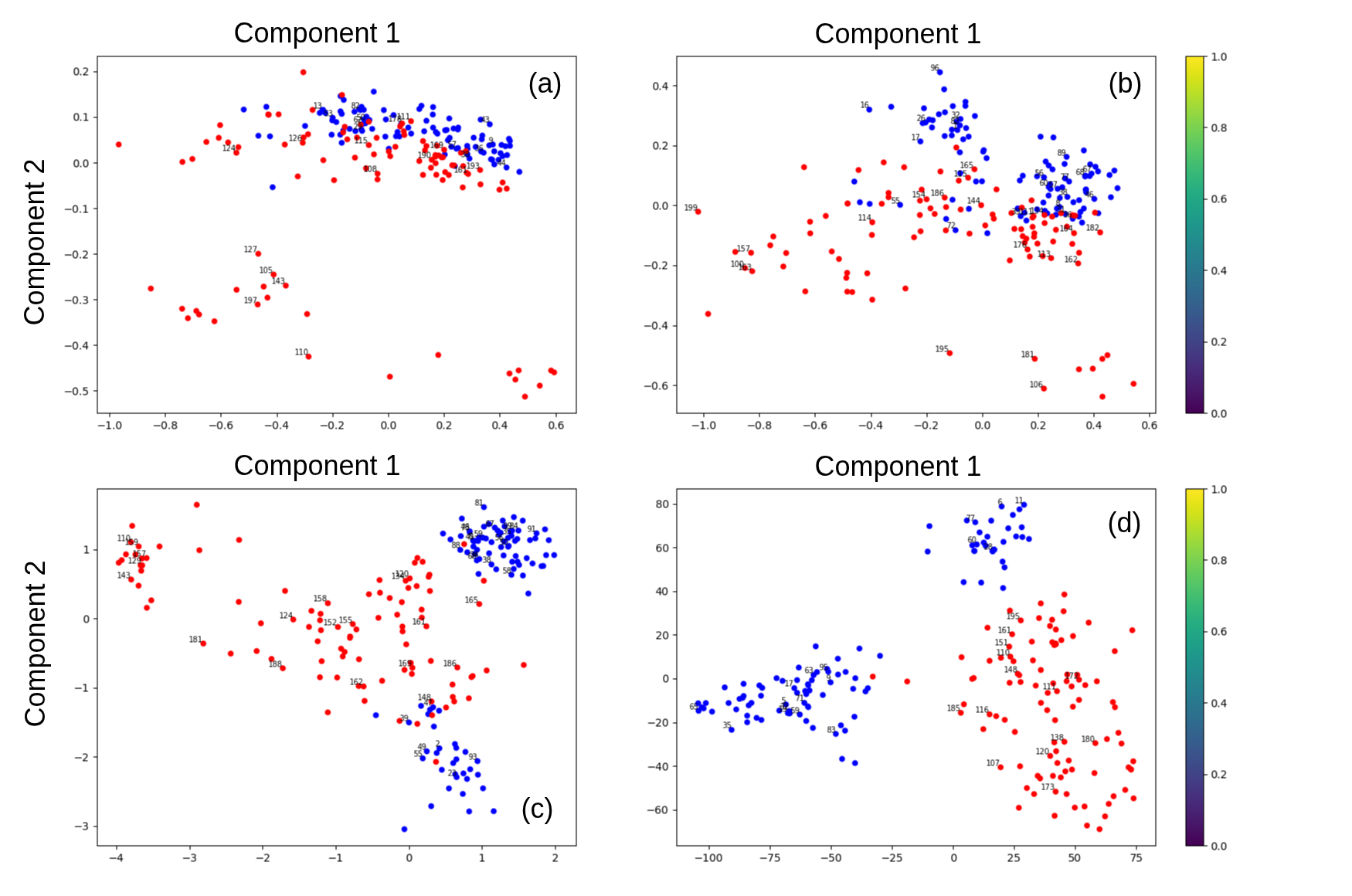}
\caption{An illustration of internal representations for toxic (\textcolor{red}{red}) and benign (\textcolor{blue}{blue}) prompts in $\inspectionModelVLMThree$: (a) the initial layer fails to differentiate benign prompts from the toxic ones, (b) by the 3rd layer respective clusters for both the benign and toxic prompts starts to emerge; (c) a clear distinction appears at the seventh layer, that (d) gradually matures and remains intact till the last layer.}
    \label{fig:miniGPT4_onlyText}
\end{figure}

\subsection{Observation for Image-based Inputs} \label{sec:supplement_imageBasedObservation}
\begin{figure}[!ht]
\centering
\includegraphics[clip, trim={0 0 5cm 0},width=1\linewidth]{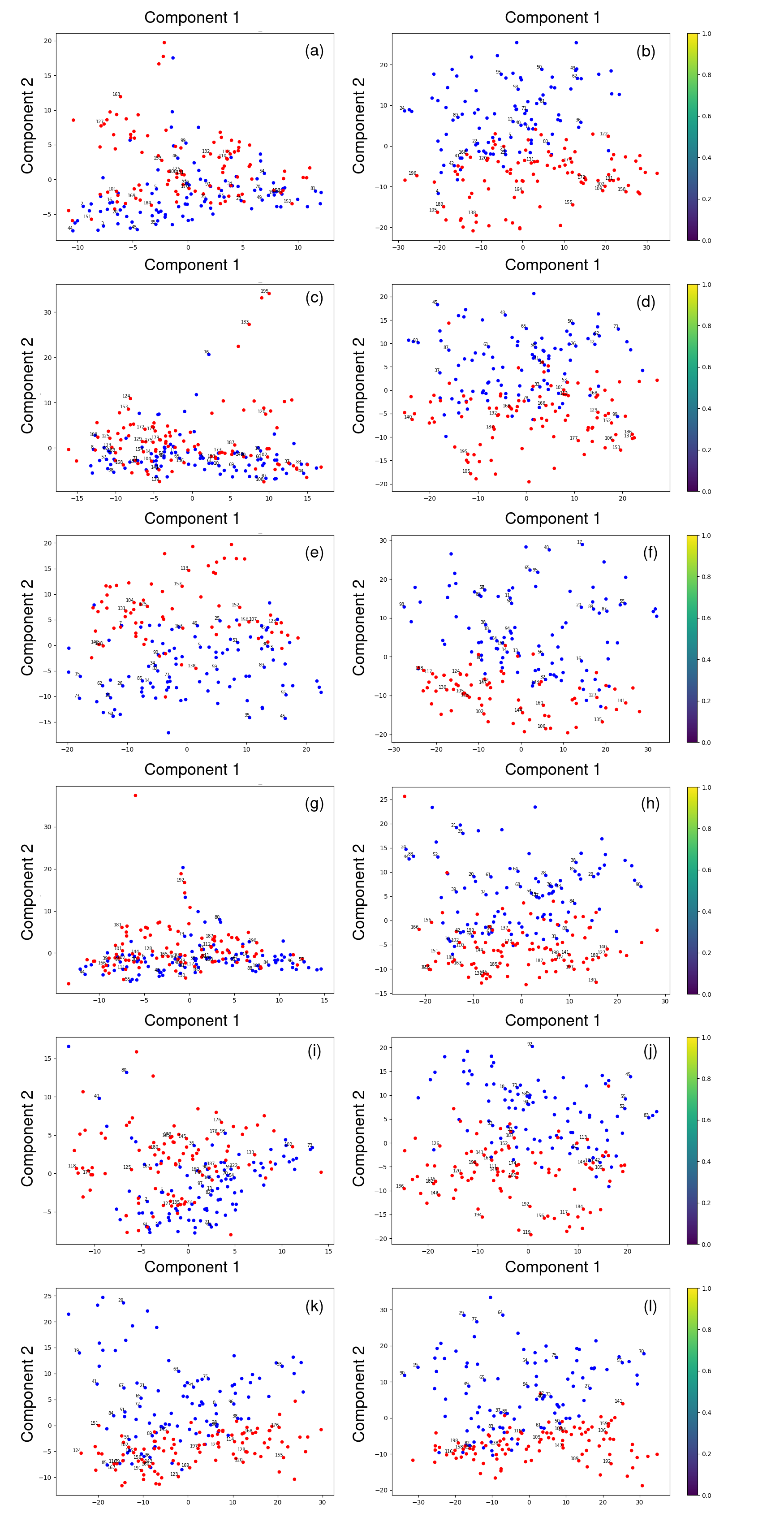}
\caption{An illustration of the internal representations of toxic (\textcolor{red}{red}) and benign (\textcolor{blue}{blue}) visual inputs for the model $\inspectionModelVLMOne$, across various targeted communities. Subfigures (a)–(b) distinguish between toxic memes targeting \textit{women} and benign images; (c)–(d) present similar contrasts for the \textit{African-American} community; (e)–(f) focus on memes directed at \textit{disabled} individuals; (g)–(h) examine toxic content related to \textit{Islam}; and (i)–(j) address toxic representations targeting the \textit{LGBTQ} community. Lastly, (k)–(l) compare \textit{general toxic} visual inputs with benign counterparts.}
    \label{fig:Llava_1p5_onlyImage}
\end{figure}

\begin{figure}[!t]
\centering
\includegraphics[clip, trim={0 0 0 0},width=1\linewidth]{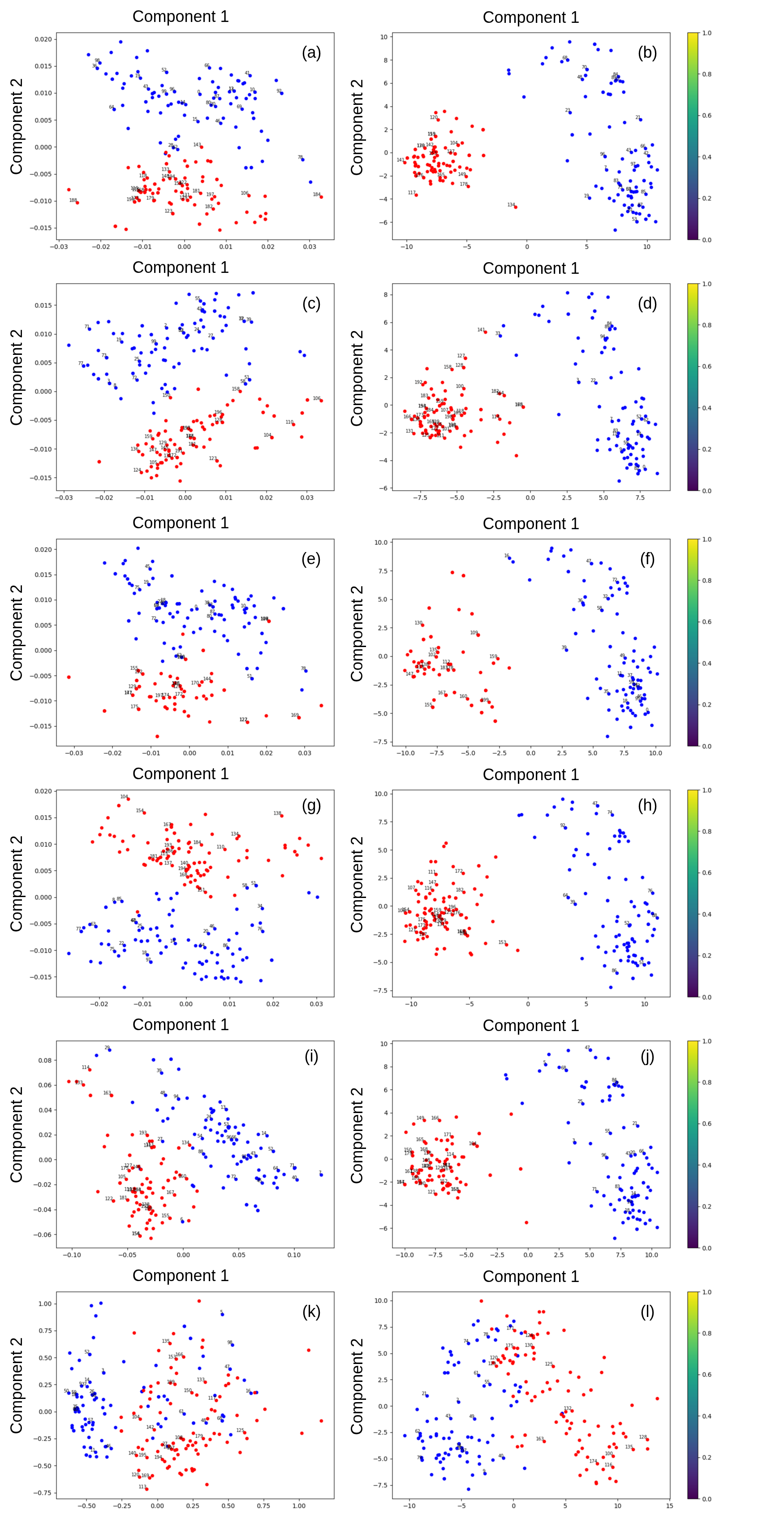}
\caption{An illustration of internal representations of toxic (\textcolor{red}{red}) and benign (\textcolor{blue}{blue}) visual inputs (only) for model $\inspectionModelVLMTwo$ across different targeted communities: (a)–(b) show the distinction between toxic memes targeting the  \text{women} and benign images, (c)–(d) illustrate a similar trend for toxic memes targeting the \text{African-American} community, (e)–(f) analyze toxic memes directed at \text{disabled} individuals, (g)–(h) examine toxic visual content related to \text{Islam}, (i)–(j) depict toxic memes targeting the \text{LGBTQ} community. Finally, (k)–(l) compare \text{general toxic} visual instances against benign images.}
\label{fig:llavaMistral_1p6_onlyImage}
\end{figure}

\begin{figure}[!ht]
\centering
\includegraphics[clip, trim={0 0 0 0},width=1\linewidth]{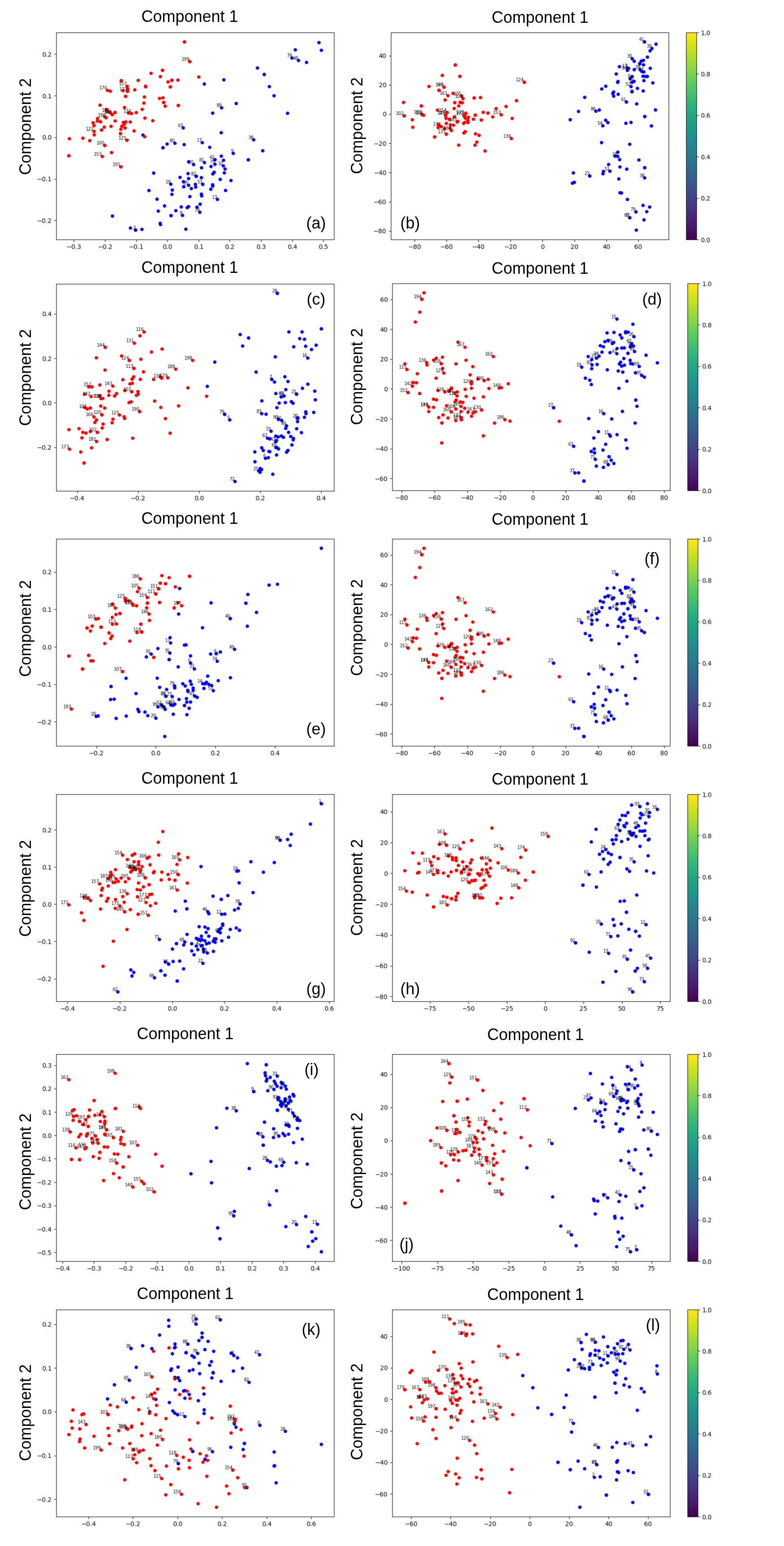}
\caption{An illustration of internal representations of toxic (\textcolor{red}{red}) and benign (\textcolor{blue}{blue}) visual inputs (only) for model $\inspectionModelVLMThree$ across different targeted communities: (a)–(b) show the distinction between toxic memes targeting the  \text{women} and benign images, (c)–(d) illustrate a similar trend for toxic memes targeting the \text{African-American} community, (e)–(f) analyze toxic memes directed at \text{disabled} individuals, (g)–(h) examine toxic visual content related to \text{Islam}, (i)–(j) depict toxic memes targeting the \text{LGBTQ} community. Finally, (k)–(l) compare \text{general toxic} visual instances against benign images.}
\label{fig:miniGPT4_1p6_onlyImage}
\end{figure}

The benign and each subcategory of toxic class, contains a total of $100$ visual instances. Each row in Figure \ref{fig:Llava_1p5_onlyImage}, Figure \ref{fig:llavaMistral_1p6_onlyImage}, and Figure \ref{fig:miniGPT4_1p6_onlyImage} presents a comparative analysis between a specific toxicity subcategory and benign examples for $\inspectionModelVLMOne$, $\inspectionModelVLMTwo$, and $\inspectionModelVLMThree$, respectively; it describes the specific layers in the model where the differentiation between toxic and benign images emerges, highlighting the stages where these distinctions become visually prominent. 

Figure \ref{fig:Llava_1p5_onlyImage} provides an analysis of benign images and toxic memes across different categories. Subfigures (a) and (b) illustrate the differentiation between toxic memes targeting the benign and the women category, with separation emerging in the 14th layer and reaching optimality at the 23rd layer. A similar trend is observed in the subfigures (c) and (d), where the distinction for the African-American black community appears later, beginning at the 19th layer and optimizing at the 22nd layer. In subfigures (e) and (f), the analysis is based on the disabled individuals, with initial differentiation observed in the 21st layer and maturing at the 24th layer. For toxic visual content related to Islam (c.f. Figure \ref{fig:Llava_1p5_onlyImage}(g) and h), the distinction emerges at the 17th layer and peaks at the 22nd layer. Subfigures (i) and (j) depict the contrast between toxic memes directed at the LGBTQ community and benign images, where an early separation is observed at the 10th layer, optimizing at the 22nd layer. Lastly, figure \ref{fig:Llava_1p5_onlyImage}.k and \ref{fig:Llava_1p5_onlyImage}.l examine the general toxic visual instances versus benign images, with the distinction appearing relatively late at the 22nd layer and reaching its optimal separation at the 23rd layer. For both general toxic and meme visuals, the separation becomes prominent between 21st to 24th layer.

Figure \ref{fig:llavaMistral_1p6_onlyImage} illustrates the separability between various sub-groups of toxic memes and benign visual inputs in $\inspectionModelVLMTwo$. Subfigures (a)–(b), (c)–(d), (e)–(f), (g)–(h), (i)–(j), and (k)–(l) correspond to memes targeting women, African-Americans, individuals with disabilities, the Islamic community, the LGBTQ community, and general toxic content, respectively. In all the mentioned sub-categories—excluding LGBTQ community and general toxic visuals, the separation between toxic and benign inputs is clear from the initial layer and remains consistently distinguishable through to the final layer. In case of LGBTQ community and the general toxic visuals, the benign and toxic input visuals are identifiable after third and 14th layer, respectively.

Interestingly, a similar pattern to that observed in $\inspectionModelVLMTwo$ emerges in $\inspectionModelVLMThree$, particularly in the visual representations across meme sub-categories when compared to benign counterparts (c.f. Figure \ref{fig:miniGPT4_1p6_onlyImage}). Subfigures (a)–(b), (c)–(d), (e)–(f), (g)–(h), (i)–(j), and (k)–(l) correspond to memes targeting women, African-Americans, individuals with disabilities, the Islamic community, the LGBTQ community, and general toxic content, respectively. For $\inspectionModelVLMThree$, across all examined sub-categories, the distinction between toxic and benign inputs is apparent from the initial layer and remains consistent throughout all the remaining layers up to the final one.

\begin{figure}[!t]
\centering
\includegraphics[clip, trim={0 1cm 0 0},width=1\linewidth]{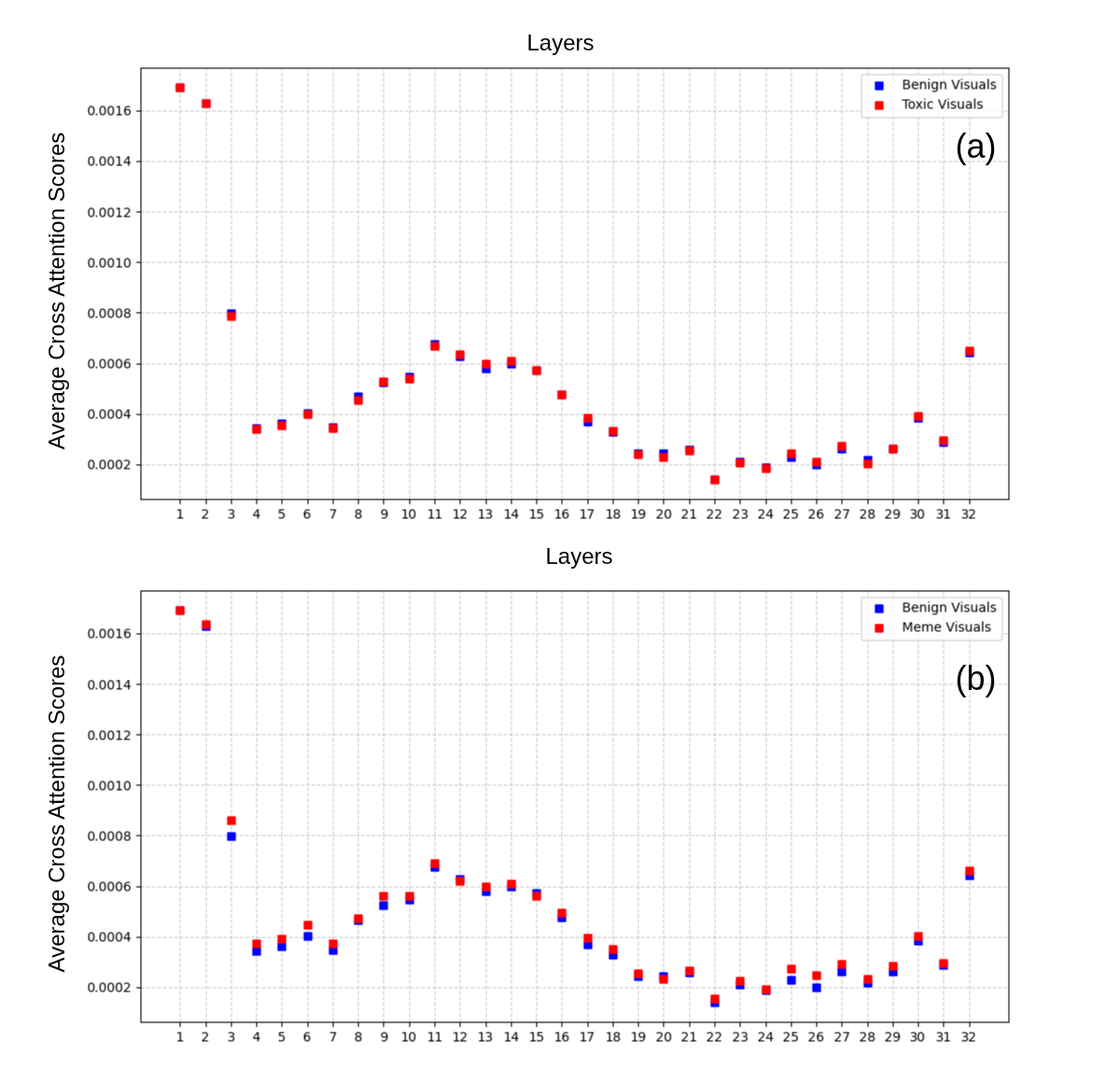}
\caption{An illustration of the influence of visual inputs on attention scores when prompted with unsafe queries for $\inspectionModelVLMOne$: (a) shows the average cross-attention scores elicited by unsafe textual prompts when paired with either benign or toxic images , (b) demonstrates the same when paired with either benign or meme images.}
\label{fig:attentionVisualization_forLLaVa1p5_V}
\end{figure}

\begin{figure}[!t]
\centering
\includegraphics[clip, trim={0 1cm 0 0},width=1\linewidth]{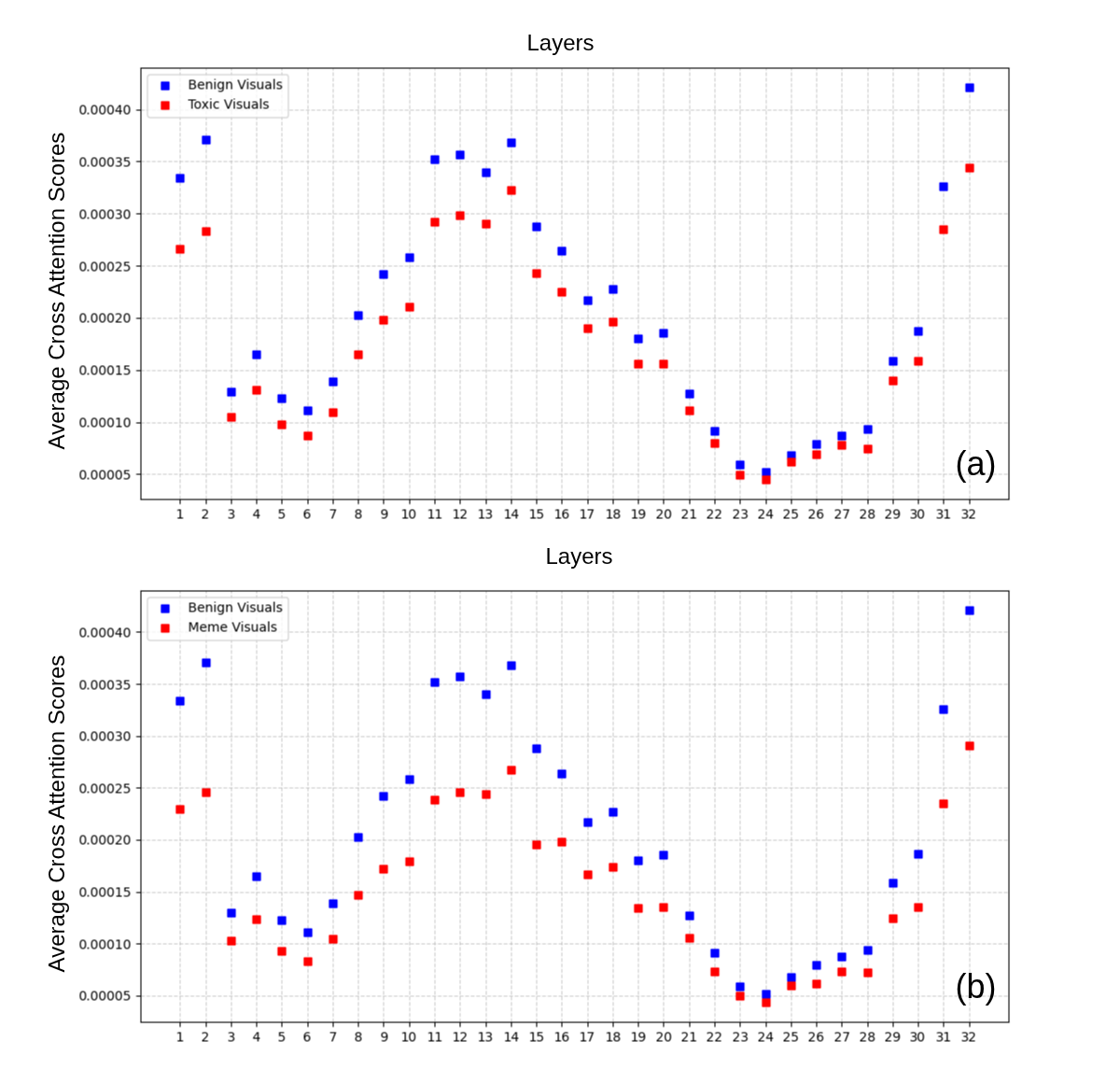}
\caption{An illustration of the influence of visual inputs on attention scores when prompted with unsafe queries for $\inspectionModelVLMTwo$: (a) shows the average cross-attention scores elicited by unsafe textual prompts when paired with either benign or toxic images , (b) demonstrates the same when paired with either benign or meme images.}
\label{fig:attentionVisualization_forLLaVa1p6_M}
\end{figure}

\subsection{Preliminary Observation on Attention Scores for Image-based Inputs in $\inspectionModelVLMOne$} \label{sec:supplement_attentionScores}

Additionally, we examined the attention scores when an unsafe input is prompted with benign, toxic, and meme-based visual content. Figure \ref{fig:attentionVisualization_forLLaVa1p5_V} and \ref{fig:attentionVisualization_forLLaVa1p6_M} demonstrates the average cross-attention scores across all layers for model, $\inspectionModelVLMOne$ and $\inspectionModelVLMTwo$ when unsafe textual prompts paired with different categories of visual inputs. In the case of $\inspectionModelVLMOne$, for toxic visuals, the average attention scores demonstrate a gradual increment over benign visuals (c.f. Figure \ref{fig:attentionVisualization_forLLaVa1p5_V}.(a)) that is consistent with the observations in Figure \ref{fig:Llava_1p5_onlyImage}. Interestingly, the average attention scores for meme-based visuals are higher (or equal) than benign visuals, starting from the initial layer (c.f. Figure \ref{fig:attentionVisualization_forLLaVa1p5_V}.(b)). Interestingly, the average cross-attention scores for the toxic and meme visuals are consistently lower than benign visuals in $\inspectionModelVLMTwo$ (c.f. Figure \ref{fig:attentionVisualization_forLLaVa1p6_M}). This observation endorses the hypothesis that memes-based visuals can be equivalently effective as toxic visuals in a successful jailbreak.

\end{document}